\documentclass{article}

\PassOptionsToPackage{numbers, compress}{natbib}
\usepackage[preprint]{neurips_2026}

\usepackage[utf8]{inputenc} 
\usepackage[T1]{fontenc}    
\usepackage{hyperref}       
\usepackage{url}            
\usepackage{booktabs}       
\usepackage{amsfonts}       
\usepackage{nicefrac}       
\usepackage{microtype}      
\usepackage{xcolor}         
\definecolor{igpositive}{RGB}{0,128,0}
\definecolor{goldanswer}{RGB}{0,80,180}
\usepackage{amsmath} 
\usepackage{algorithm} 
\usepackage{algorithmic} 
\usepackage{colortbl} 
\usepackage{multirow} 
\usepackage{graphicx,lipsum,afterpage,subcaption}
\usepackage{makecell}

\title{IG-Search: Step-Level Information Gain Rewards for Search-Augmented Reasoning}

\author{%
  Zihan Liang\thanks{Equal Contribution}, \quad 
  Yufei Ma\footnotemark[1], \quad
  Ben Chen\footnotemark[1] \thanks{Corresponding author}, \quad Zhipeng Qian, \quad Huangyu Dai \\
  \textbf{Lingtao Mao}, \quad \textbf{Xuxin Zhang}, \quad \textbf{Chenyi Lei}, \quad \textbf{Wenwu Ou} \\
  \\
  Kuaishou Technology \\
  \texttt{benchen4395@gmail.com}
}

\begin{document}

\maketitle

\begin{abstract}
Reinforcement learning has emerged as an effective paradigm for training large language models to perform search-augmented reasoning. 
However, existing approaches rely on trajectory-level rewards that cannot distinguish precise search queries from vague or redundant ones within a rollout group, and collapse to a near-zero gradient signal whenever every sampled trajectory fails.
In this paper, we propose \textbf{IG-Search}, a reinforcement learning framework that introduces a step-level reward based on Information Gain (IG). 
For each search step, IG measures how much the retrieved documents improve the model's confidence in the gold answer relative to a counterfactual baseline of random documents, thereby reflecting the effectiveness of the underlying search query.
This signal is fed back to the corresponding search-query tokens via per-token advantage modulation in GRPO, enabling fine-grained, step-level credit assignment within a rollout.
Unlike prior step-level methods that require either externally annotated intermediate supervision or shared environment states across trajectories, IG-Search derives its signals from the policy's own generation probabilities, requiring no intermediate annotations beyond standard question-answer pairs.
Experiments on seven single-hop and multi-hop QA benchmarks demonstrate that IG-Search achieves an average EM of 0.430 with Qwen2.5-3B, outperforming the strongest trajectory-level baseline~(MR-Search) by 1.6~points and the step-level method GiGPO by 0.9~points on average across benchmarks, with particularly pronounced gains on multi-hop reasoning tasks. 
Despite introducing a dense step-level signal, IG-Search adds only ${\sim}6.4\%$ to per-step training wall-clock time over the trajectory-level baseline and leaves inference latency unchanged, while still providing a meaningful gradient signal even when every sampled trajectory answers incorrectly.

\end{abstract}

\section{Introduction}
\label{sec:intro}

\par
Large language models (LLMs) have shown impressive abilities in language understanding, planning, and problem solving~\citep{openai2023gpt4,touvron2023llama,qwen2025}. 
Recent advances demonstrate that reinforcement learning (RL) further enhances LLMs' reasoning capabilities~\citep{deepseek2025r1,openai2024openaio1card}, especially in complex tasks such as mathematical problem solving~\citep{shao2024deepseekmath} and code generation~\citep{guo2024deepseekcoder}. 
However, the knowledge encoded in LLMs is inherently bounded by their training corpora, limiting their reasoning performance on tasks that require up-to-date or specialized factual information~\citep{jiang2023active,shao2023enhancing}.

\par
A common strategy to address this limitation is retrieval-augmented generation (RAG), which equips LLMs with retrieval tools to access external knowledge bases during reasoning~\citep{lewis2020naiverag,jiang2023activerag,ram2023incontext}. 
While early RAG approaches rely on supervised fine-tuning (SFT) to teach models when and how to search~\citep{asai2024selfrag,shi2023replug,yan2024corrective}, recent work draws inspiration from RL-based post-training~\citep{deepseek2025r1} and explores RL for retrieval-augmented reasoning, forming a \emph{search-during-think} paradigm where the model autonomously invokes retrieval tools within its reasoning trajectories~\citep{jin2025searchr1,chen2025research,song2025r1searcher}. 
Building on this paradigm, AutoRefine~\citep{shi2025autorefine} introduces an explicit knowledge refinement step after each retrieval call and incorporates a retrieval-specific reward alongside the answer correctness reward via Group Relative Policy Optimization (GRPO)~\citep{shao2024deepseekmath}, achieving strong performance on both single-hop and multi-hop question answering scenarios. 
MR-Search~\citep{xiao2025mrsearch} further introduces cross-episode self-reflection, but assigns credit at the episode level rather than to individual search steps.

\par
Despite these promising results, we observe that reward signals in existing retrieval-augmented reasoning methods operate at trajectory level. 
In GRPO, a group of trajectories is sampled for each question, and each trajectory's advantage is determined by its overall reward relative to the group. 
This design introduces two structural issues that limit fine-grained learning of search behaviors.

\begin{figure*}
    \includegraphics[width=\textwidth]{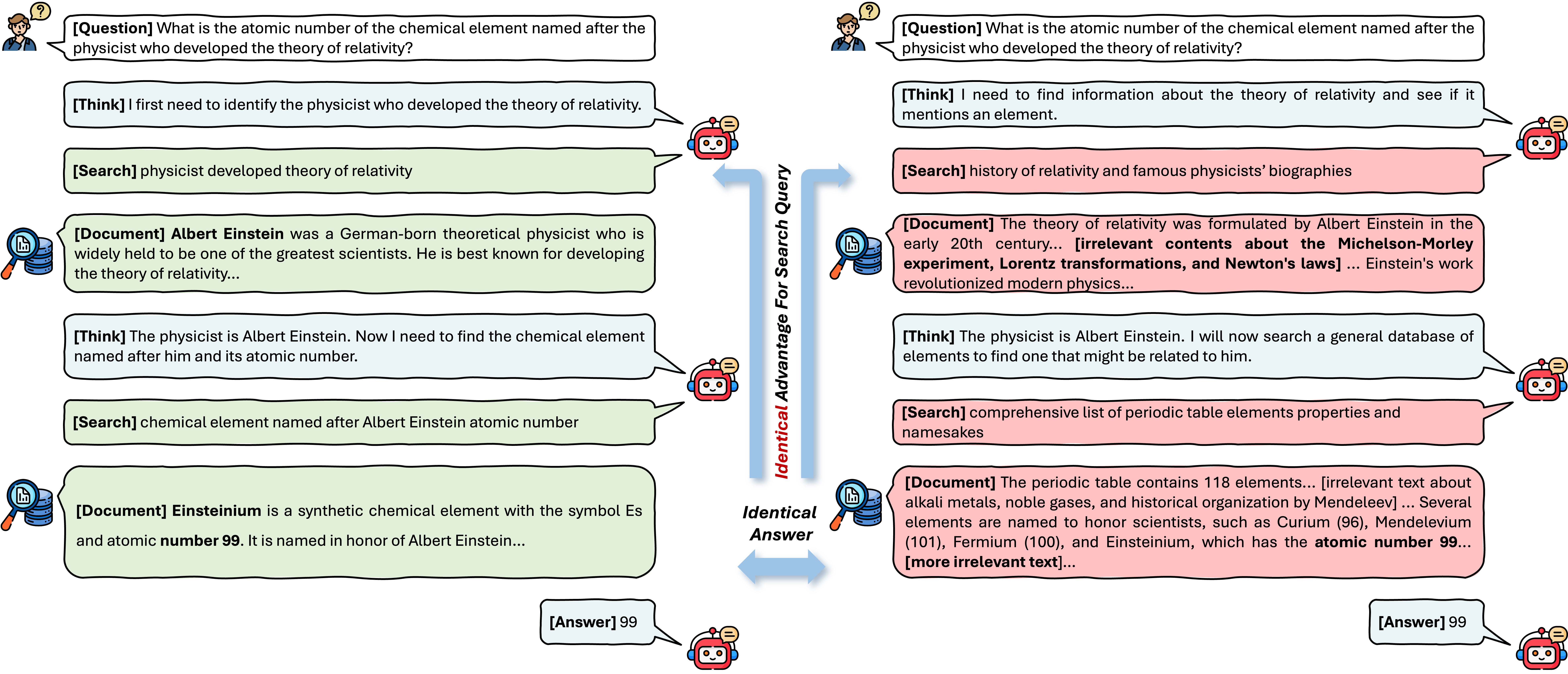}
    \caption{Trajectory-level rewards obscure step-level query quality. Despite the left trajectory executing a precise search and the right one retrieving irrelevant contents, both receive identical rewards for producing the same final answer.
    }
    \label{fig:inability}
\end{figure*}

\textbf{Inability to distinguish query quality within a rollout group.} 
As illustrated in Figure~\ref{fig:inability}, for a multi-hop question, one trajectory may generate a precise query that directly retrieves the answer-bearing document, while another may issue a vague query that returns mostly irrelevant content. 
If both trajectories produce the same final answer, they receive identical reward signals, leaving the model unable to learn which query formulation is more effective. 
This problem is particularly severe in multi-hop scenarios, where a suboptimal query at an early step can derail the entire reasoning chain.

\par
\textbf{Vanishing signal in the all-failure scenario.} When all trajectories fail to produce a correct answer, the trajectory-level advantages become approximately zero, yielding nearly no gradient signal. 
This scenario is common during early training and on difficult multi-hop questions, exactly when effective supervision is most needed.

\par
To address these limitations, we propose \textbf{IG-Search}, an RL framework that rewards each search step according to the Information Gain (IG) of its retrieval. 
The core idea is simple: given a fixed retriever, the quality of a search query is directly reflected in the documents it retrieves, so we can evaluate each query by measuring the informational value of those documents.
Concretely, for each search step, IG quantifies how much the actual retrieved documents improve the model's confidence in the gold answer relative to a counterfactual baseline of random documents. 
This signal is then propagated back to the corresponding query tokens through per-token advantage modulation within GRPO, so that different search steps within the same trajectory receive gradients proportional to their informational contribution, rather than sharing a single trajectory-level advantage. 
Crucially, because IG is computed independently of whether the final answer is correct, the step-level signal remains informative even when all trajectories fail. 
We further develop stabilization mechanisms to ensure robust training, detailed in \S\ref{sec:stab}.

\par
Prior works have also explored step-level supervision.
IGPO~\citep{wang2026igpo} defines a turn-level reward as the temporal difference in gold-answer log-probability between consecutive turns, but this signal conflates reasoning, querying, and retrieval within each turn rather than isolating query quality.
StepSearch~\citep{li2025stepsearch} and GiGPO~\citep{feng2025gigpo} rely respectively on GPT-4o-generated sub-question annotations and on state recurrence across trajectories.
IG-Search requires none of these: it derives a query-grounded signal via counterfactual comparison, using only standard QA pairs; we discuss these methods in detail in Appendix~\ref{sec:related_work}.

\par
We evaluate IG-Search on seven QA benchmarks spanning single-hop (NQ, TriviaQA, PopQA) and multi-hop (HotpotQA, 2WikiMultihopQA, Musique, Bamboogle) settings. 
With Qwen2.5-3B as the base model, IG-Search achieves an average Exact Match of 0.430, outperforming the strongest trajectory-level baseline (MR-Search) by 1.6 points and step-level method GiGPO by 0.9 points on average across benchmarks, with particularly pronounced gains on multi-hop reasoning tasks. 
Experiments with Qwen2.5-7B further confirm the generalizability of our method, yielding consistent improvements across all benchmarks.

\par
Our contributions are summarized as follows:
\begin{itemize}
    \item We propose IG-Search, which introduces a step-level IG reward that measures the informational contribution of each search step. IG-Search can be viewed as a \emph{retrieval-grounded process reward}: it derives signal from the policy's own generation probabilities conditioned on the gold answer, requiring neither a separately trained reward model nor intermediate annotations beyond standard question-answer pairs.
    \item We develop per-token advantage modulation within GRPO so that different search steps receive gradient signals proportional to their informational contribution, together with stabilization mechanisms that prevent degenerate behaviors. Notably, the resulting signal remains informative even when all sampled trajectories answer incorrectly.
    \item Extensive experiments on diverse QA benchmarks demonstrate that IG-Search achieves the strongest results among open-source models of comparable size at both the 3B and 7B scales, and comprehensive ablations validate the necessity of each proposed component.
\end{itemize}

\section{Method}
\label{sec:method}

\par
We introduce IG-Search, a reinforcement learning framework that provides step-level IG rewards for retrieval-augmented reasoning.
We first review the necessary background (\S\ref{sec:prelim}), then present the IG formulation (\S\ref{sec:ig}), describe how IG modulates per-token advantages in GRPO (\S\ref{sec:advmod}), and detail the stabilization mechanisms that ensure robust training (\S\ref{sec:stab}). 
The overall framework is illustrated in Figure~\ref{fig:method}, and the training pipeline is summarized in Appendix~\ref{app:training_algorithm}.

\begin{figure*}[t]
    \centering
    \includegraphics[width=\textwidth]{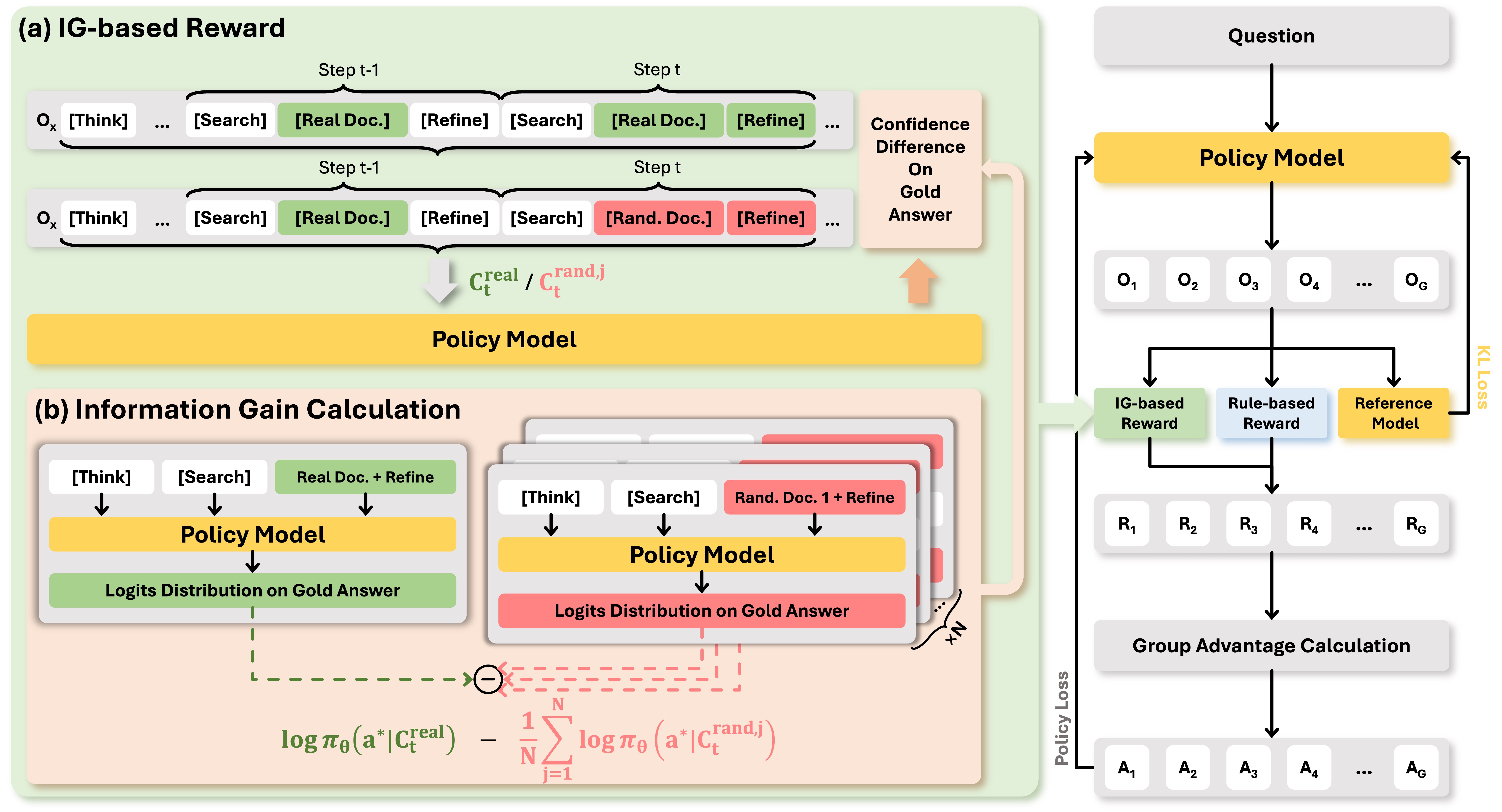}
    \caption{Overview of IG-Search. \textbf{Right:} The policy model generates $G$ trajectories per question, each scored by rule-based rewards and step-level IG rewards to compute group advantages. \textbf{Left:} IG computation for search step $t$: the log-probability of the gold answer under the real retrieval context minus the average under $N$ randomly sampled retrieval contexts.}
    \label{fig:method}
\end{figure*}

\subsection{Preliminaries}
\label{sec:prelim}
 
\paragraph{Trajectory Generation.}
Following recent work on retrieval-augmented reasoning~\citep{jin2025searchr1, shi2025autorefine}, policy model $\pi_\theta$ generates reasoning trajectories by interacting with an external retriever $\mathcal{E}$. 
Each trajectory $o = (\tau_1, \tau_2, \ldots, \tau_T)$ consists of a sequence of structured actions: \texttt{<think>} for reasoning, \texttt{<search>} for querying the search engine, \texttt{<documents>} for the content returned by the search engine, \texttt{<refine>} for distilling (by the policy) key evidence from the retrieved documents into a concise summary, and \texttt{<answer>} for producing the final answer. 
The number of search cycles is not predefined but determined autonomously by the model. For each search step $t$, we denote the generated query as $\mathbf{q}_t$, the retrieved documents as $\mathbf{d}_t = \mathcal{E}(\mathbf{q}_t)$, and the refinement output as $\mathbf{r}_t$.

\par
\paragraph{GRPO.}
We adopt GRPO~\citep{shao2024deepseekmath} as the policy optimization algorithm. For each question $q$, a group of $G$ trajectories $\{o_1, \ldots, o_G\}$ is sampled from the current policy $\pi_{\theta_{\text{old}}}$. Each trajectory $o_i$ receives a scalar reward $R_i$, and the advantage is computed via group-level normalization:
\begin{equation}
\footnotesize
\hat{A}_i = \frac{R_i - \text{mean}(\{R_j\}_{j=1}^G)}{\text{std}(\{R_j\}_{j=1}^G)}.
\label{eq:grpo_adv}
\end{equation}
In GRPO, this advantage $\hat{A}_i$ is shared across all tokens in trajectory $o_i$.

\par
\paragraph{Baseline Rewards.}
Following AutoRefine~\citep{shi2025autorefine}, we inherit a trajectory-level reward $R_i$ consisting of two components: (1) $r_{\text{F1}}$, F1 score between the predicted and gold answers; and (2) $r_{\text{ret}}$, a retrieval-specific reward that evaluates whether the \texttt{<refine>} blocks contain the gold answer.
IG-Search retains both reward components and introduces the step-level IG signal as an additional per-token advantage modulation, as detailed in \S\ref{sec:advmod}.

\subsection{Step-Level Information Gain}
\label{sec:ig}

\par
The central contribution of IG-Search is a step-level reward that directly measures the informational value of each search step. 
We define IG as the difference in the model's confidence in generating the gold answer when conditioned on the real retrieved documents relative to random documents.

\par
\paragraph{IG Definition.}
For search step $t$ in a trajectory, let $\mathcal{C}_t^{\text{real}}$ denote the full context up to and including step $t$, with all preceding steps retaining their original retrieved documents and refinements, i.e., $\mathcal{C}_t^{\text{real}} = (\tau_{<t},\, \mathbf{q}_t,\, \mathbf{d}_t,\, \mathbf{r}_t)$,
where $\tau_{<t}$ denotes all preceding steps in their entirety, and $\mathbf{q}_t$, $\mathbf{d}_t$, $\mathbf{r}_t$ are the query, retrieved documents, and refinement at the current step $t$.
We construct $N$ counterfactual contexts $\{\mathcal{C}_t^{\text{rand},j}\}_{j=1}^N$ by replacing only the (documents, refinement) pair at step $t$ with pairs drawn from randomly selected search steps of other questions in the same training batch, which ensures a consistent document-length distribution without additional data loading overhead.

The IG 
\footnote{Strictly speaking, Eq.~\ref{eq:ig} is a log-likelihood ratio comparing the real retrieval context against an empirical average over $N$ random ones, evaluated at the gold answer $a^*$. It is not a Shannon information gain but shares the same sign convention: positive values indicate that the real documents raise the model's confidence in $a^*$ relative to uninformative retrievals. We use the term ``Information Gain'' for this intuitive reading; all claims in this paper refer to this pointwise quantity.}
for step $t$ is then defined as:
\begin{equation}
\footnotesize
\text{IG}_t = \log \pi_{\theta_{\text{old}}}(a^* \mid \mathcal{C}_t^{\text{real}}) - \frac{1}{N}\sum_{j=1}^{N} \log \pi_{\theta_{\text{old}}}(a^* \mid \mathcal{C}_t^{\text{rand},j}),
\label{eq:ig}
\end{equation}
where $a^*$ denotes the gold answer, and $\pi_{\theta_{\text{old}}}$ is the rollout policy used to sample the trajectories.

The real-context term $\log \pi_{\theta_{\text{old}}}(a^* \mid \mathcal{C}_t^{\text{real}})$ and the $N$ counterfactual terms can be computed highly efficiently. By stacking the real context and the $N$ random contexts into a single batch, we obtain the log-probabilities of the gold answer for all $N{+}1$ variants in a single forward pass. Crucially, this batched pass scores only the short gold-answer tokens (typically a few subwords) rather than generating new sequences; the resulting overhead is detailed in Appendix~\ref{app:overhead}.
When multiple valid answers exist, we compute the log-probability for each variant and average the values.
We provide a concrete example of the IG computation on a two-hop question in Appendix~\ref{app:ig_example}, illustrating how IG captures the marginal informational contribution of each successive search step.
The log-probability of the gold answer is computed as the length-normalized sum of per-token log-probabilities:
\begin{equation}
\footnotesize
\log \pi_{\theta_{\text{old}}}(a^* \mid \mathcal{C}) = \frac{1}{|a^*|}\sum_{k=1}^{|a^*|} \log \pi_{\theta_{\text{old}}}(a^*_k \mid \mathcal{C}, a^*_{<k}).
\label{eq:logp}
\end{equation}
Because Eq.~\ref{eq:logp} length-normalizes by $|a^*|$, all subsequent IG values (including the thresholds $\delta$ and $\eta$) are reported in \emph{nats per answer token}, making them comparable across answers of different length.

\par
\paragraph{Why against Random Documents?}
A natural alternative is to compare against an empty context. 
However, real and empty contexts differ not only in informativeness but also in length and positional structure, introducing a systematic bias unrelated to retrieval quality. 
Drawing counterfactual documents from other questions in the same batch matches the length and structural characteristics of the real context while removing its topical relevance, so that $\mathrm{IG}_t$ isolates the informational contribution of retrieval. 
Figure~\ref{fig:baseline_design} confirms that the empty baseline performs worst, with its degradation mainly concentrated on multi-hop scenarios.

\subsection{Advantage Modulation with IG}
\label{sec:advmod}

\par
In GRPO, all tokens in a trajectory share identical $\hat{A}_i$ (Eq.~\ref{eq:grpo_adv}). 
IG-Search breaks this uniformity by modulating the advantage of query tokens based on the IG of the corresponding search step.

\par
Concretely, let $\mathcal{Q}_t$ denote the set of token positions that belong to the search query at step $t$ (i.e., tokens within $\texttt{<search>}\ldots\texttt{</search>}$). 
The modulated advantage for each token at position $p$ in trajectory $o_i$ is:
\begin{equation}
\footnotesize
\tilde{A}_{i,p} =
\begin{cases}
\hat{A}_i + \alpha \cdot \dfrac{\widetilde{\text{IG}}_t}{|\mathcal{Q}_t|}, & \text{if } p \in \mathcal{Q}_t, \\[6pt]
\hat{A}_i, & \text{otherwise},
\end{cases}
\label{eq:adv_mod}
\end{equation}
where $\widetilde{\text{IG}}_t$ is the processed IG value (detailed in \S\ref{sec:stab}), $|\mathcal{Q}_t|$ denotes the number of tokens in the search query, and $\alpha$ is a weighting coefficient.

\par
The division by $|\mathcal{Q}_t|$ ensures that a search step's total contribution to the trajectory loss equals $\alpha \cdot \widetilde{\text{IG}}_t$ and is therefore decoupled from the query length, which is essential to prevent the reward-hacking behavior discussed in \S\ref{sec:stab}.

\par
We modulate only the query tokens instead of all tokens in the trajectory because IG directly reflects the quality of the documents retrieved by the fixed retriever, which is determined by the search query.
This formulation has two vital properties. 
First, in a single trajectory, different search steps receive different gradient signals proportional to their informational contribution, enabling the model to learn which query formulations are more effective. 
Second, even when all trajectories answer incorrectly, the IG term $\alpha \cdot \widetilde{\text{IG}}_t / |\mathcal{Q}_t|$ still ensures nonzero gradients to query tokens, allowing the model to learn from the relative quality of different search steps across otherwise indistinguishable trajectories.

\subsection{Stabilization Mechanisms}
\label{sec:stab}

\par
As observed in prior work on policy optimization and language model alignment \citep{schulman2017proximal, rafailov2024dpo, jiang2024howcanweknowwhatLMknow}, metrics derived from the difference or ratio of raw LLM log-probabilities inherently exhibit high variance and heavy-tailed distributions. 
If used directly, these extreme values and estimation noise can destabilize training or induce degenerate behaviors, similar to the unbounded implicit rewards in preference learning. 
To ensure robust optimization, we process the raw $\text{IG}_t$ into $\widetilde{\text{IG}}_t$ through a pipeline of stabilization mechanisms.

\par
\paragraph{Dead Zone Filtering.}
For questions the model can answer from parametric knowledge, IG does not faithfully reflect query quality: relevant documents add little new information and yield near-zero IG, while real documents can even introduce competing facts that slightly dilute the parametric prior, producing small negative IG for well-formed queries (see Case~3 in \S\ref{sec:deadzone_cases}).
In both cases the value of IG is driven by the model's parametric knowledge rather than by the effectiveness of the query.
We therefore apply a dead zone: if $|\text{IG}_t| < \delta$, we set $\text{IG}_t = 0$. We use $\delta = 0.5$ by default.

\paragraph{Asymmetric Negative Scaling.}
A negative IG indicates that the actual retrieved documents are less helpful than random ones, suggesting a poor search query. 
However, we observe that applying full-strength penalties for negative IG can cause the model to collapse to a degenerate strategy of never issuing search queries, thereby avoiding all negative signals. 
To prevent this, we scale negative IG values by a factor $\lambda < 1$:
\begin{equation}
\footnotesize
\text{IG}_t \leftarrow
\begin{cases}
\text{IG}_t, & \text{if } \text{IG}_t \geq 0, \\
\lambda \cdot \text{IG}_t, & \text{if } \text{IG}_t < 0.
\end{cases}
\label{eq:asymmetric}
\end{equation}
By default, we set $\lambda = 0.1$. This asymmetry ensures that the optimal strategy is to search \emph{and search well}, rather than to avoid searching entirely.

\par
We retain a small nonzero $\lambda$ rather than setting it to zero, because fully ignoring negative signals leaves poorly formulated queries gradient-wise indistinguishable from non-query tokens; the sensitivity analysis in Figure~\ref{fig:hyperparam} confirms that $\lambda{=}0$ underperforms $\lambda{=}0.1$.

\paragraph{Soft Clipping.}
Extreme IG values can dominate the reward signal and cause gradient instability. We apply a logarithmic soft clipping for values exceeding a threshold $\eta$:
\begin{equation}
\footnotesize
\text{IG}_t \leftarrow
\begin{cases}
\text{IG}_t, & \text{if } |\text{IG}_t| \leq \eta, \\
\text{sign}(\text{IG}_t) \cdot \bigl(\eta + \log(1 + |\text{IG}_t| - \eta)\bigr), & \text{otherwise}.
\end{cases}
\label{eq:softclip}
\end{equation}
This preserves the ordering of extreme values while substantially reducing their magnitude. We use $\eta = 3.0$ by default.
 
\paragraph{Query Length Normalization.}
As defined in Eq.~\ref{eq:adv_mod}, the IG bonus is divided by the query length $|\mathcal{Q}_t|$. Without this normalization, the model can exploit a reward hacking strategy by generating unnecessarily long queries: since each query token receives the IG bonus, longer queries accumulate larger total bonuses when the IG is positive.

\par
The three mechanisms are applied in a fixed order to produce $\widetilde{\text{IG}}_t$: (i)~dead zone filtering zeroes out values with $|\text{IG}_t| < \delta$; (ii)~asymmetric negative scaling compresses the surviving negative values by $\lambda$; (iii)~soft clipping bounds the remaining extreme (predominantly positive) values.
Empirically, the raw negative IG values surviving the dead zone rarely exceed $-3$ nats, so after scaling by $\lambda{=}0.1$ they lie approximately within $[-0.3, -0.05]$, well below the soft-clipping threshold~$\eta{=}3$.
We visualize the raw IG distribution at three training checkpoints in Appendix~\ref{app:ig_dist}, confirming that each mechanism addresses a substantial and distinct fraction of the search steps.

\section{Experiments}
\label{sec:exp}

\par
We aim to answer the following research questions:

\textbf{RQ1}: Does IG-Search improve QA accuracy compared to existing retrieval-augmented reasoning methods?

\textbf{RQ2}: How essential is each component of the IG mechanism?

\textbf{RQ3}: How does the choice of counterfactual baseline affect IG quality?

\textbf{RQ4}: How sensitive is IG-Search to its hyperparameters?

\textbf{RQ5}: Does IG-Search scale to larger models?

\textbf{RQ6}: What search behaviors does IG-Search learn during training, and does it provide meaningful learning signal even when all sampled trajectories fail?

\subsection{Experimental Setup}
\label{sec:setup}
 
\paragraph{Datasets.}
We evaluate on seven diverse QA benchmarks: three single-hop datasets, Natural Questions (NQ)~\citep{kwiatkowski2019naturalquestions}, TriviaQA~\citep{joshi2017triviaqa}, and PopQA~\citep{mallen2023popqa}, along with four multi-hop datasets, HotpotQA~\citep{yang2018hotpotqa}, 2WikiMultihopQA (2Wiki)~\citep{ho20202wiki}, Musique~\citep{trivedi2022musique}, and Bamboogle~\citep{press2023bamboogle}. 
Following prior work~\citep{jin2025searchr1,shi2025autorefine}, we train on a combined training set from NQ and HotpotQA and evaluate on all seven benchmarks. 
Exact Match (EM) accuracy serves as the evaluation metric. 
We normalize both predicted and ground-truth answers by lowercasing, removing punctuation, and stripping articles before comparison, and treat the prediction as correct if it exactly matches any of the ground-truth aliases.
Full dataset statistics and split information are provided in Appendix~\ref{app:datasets}.
 
\paragraph{Baselines.}

\par
We compare against methods spanning three categories:
(1)~\emph{without retrieval}: direct generation, supervised fine-tuning (SFT), and R1-style RL training~\citep{deepseek2025r1};
(2)~\emph{single-hop retrieval}: Naive RAG~\citep{lewis2020naiverag};
(3)~\emph{multi-hop retrieval}: Search-o1~\citep{li2025searcho1}, IRCoT~\citep{trivedi2023ircot}, Search-R1~\citep{jin2025searchr1}, ReSearch~\citep{chen2025research}, StepSearch~\citep{li2025stepsearch}, AutoRefine~\citep{shi2025autorefine}, MR-Search~\citep{xiao2025mrsearch}, and GiGPO~\citep{feng2025gigpo}, a step-level method that constructs step-level advantage groups by retroactively identifying repeated environment states across trajectories within the same rollout group.
All methods share the same retrieval corpus (December 2018 Wikipedia dump), retriever (E5-base-v2), and retrieval depth (top-3 documents per query). We note that StepSearch differs from other methods in both training data and optimization: it is trained with PPO on MuSiQue using GPT-4o-augmented sub-question decompositions, whereas IG-Search and the other baselines use GRPO on NQ+HotpotQA. As a result, StepSearch reports results only on the multi-hop benchmarks.
Baseline numbers are either reproduced under our setup or taken from the original papers when the evaluation protocol matches ours.

\paragraph{Implementation Details.}
We train on 8 NVIDIA H800 GPUs using the veRL framework~\citep{Sheng2025hybridflow}. 
The primary experiments use Qwen2.5-3B~\citep{qwen2025} in both base and instruct variants; scaling experiments with Qwen2.5-7B are presented in \S\ref{sec:model_size}. 
The retrieval engine returns the top 3 documents per query. 
We sample $G{=}5$ trajectories per question with temperature 1.0. 
Each rollout is allowed at most $T_{\max}=5$ search calls, after which the model is forced to produce a final answer.
For IG computation, we construct $N{=}3$ counterfactual contexts per search step using randomly sampled (documents, refinement) pairs from the training batch. 
The IG hyperparameters are: $\alpha{=}0.3$, $\delta{=}0.5$, $\eta{=}3.0$, $\lambda{=}0.1$. 
We use a single set of hyperparameters throughout all experiments without dataset-specific tuning; their robustness is analyzed in Section~\ref{sec:hyperparam}.
A complete hyperparameter listing is provided in Appendix~\ref{app:hparams}.
Training runs for 200 steps with a learning rate of $1 \times 10^{-6}$ and KL coefficient $\beta{=}0.001$.
Since the IG computation scores only the short gold-answer tokens rather than generating full sequences, and leverages a single batched forward pass with shared prefix computation across the variants, its overhead is highly efficient: introducing only an approximately 6.4\% increase in per-step wall-clock time relative to AutoRefine under identical hardware and batch size. A detailed breakdown is provided in Appendix~\ref{app:overhead}.

\subsection{Overall Performance (RQ1)}
\label{sec:main_results}

\par
Table~\ref{tab:main} presents the performance comparison on Qwen2.5-3B. 

\par
Both IG-Search variants (Base and Instruct) achieve an average EM of 0.430, the highest among all compared methods.
Against the strongest trajectory-level baseline MR-Search (0.414, also Base), IG-Search-Base improves by 1.6 points, with pronounced gains on multi-hop benchmarks (HotpotQA $+1.7$, 2Wiki $+1.4$, Musique $+1.4$), and also surpasses AutoRefine-Base (0.405) by 2.5 points on average.

\par
Because GiGPO is trained exclusively on the Instruct variant, we compare it against IG-Search-Instruct, which improves over GiGPO by 0.9 points on the seven-benchmark average, with particularly pronounced gains on HotpotQA ($+6.2$), 2Wiki ($+7.0$), and Musique ($+6.5$). GiGPO leads only on Bamboogle (0.641 vs.\ 0.424, on 125 questions), while IG-Search outperforms GiGPO on the remaining six benchmarks.

\par
StepSearch achieves competitive multi-hop results (e.g., Musique: 0.181 for the base variant), but does so by relying on GPT-4o-generated sub-question decompositions with reference documents and search keywords. 
Using only standard question-answer pairs, IG-Search matches or exceeds StepSearch on every overlapping benchmark (HotpotQA 0.436 vs.\ 0.329, 2Wiki 0.415 vs.\ 0.339, Musique 0.179 vs.\ 0.181, Bamboogle 0.390 vs.\ 0.328), showing that the policy's own generation confidence is a sufficient and more practical proxy for step-level supervision.

\par
IG-Search-Instruct achieves the same average EM of 0.430, outperforming AutoRefine-Instruct (0.396) by 3.4 points.
The two IG-Search variants exhibit complementary strengths: the base variant performs better on single-hop benchmarks (e.g., NQ: 0.476 vs.\ 0.450), while the instruct variant excels on harder multi-hop benchmarks (2Wiki: 0.440 vs.\ 0.415; Musique: 0.191 vs.\ 0.179; Bamboogle: 0.424 vs.\ 0.390).
This suggests that the instruction-following prior of the instruct model benefits complex multi-hop reasoning, while the base model may benefit from fewer inductive biases during RL exploration.
We verify the statistical significance of these improvements with a three-seed rerun in Appendix~\ref{app:statistical}.

\begin{table}[h]
\centering
\caption{(RQ1) Exact Match accuracy of IG-Search versus baseline methods on seven QA benchmarks with Qwen2.5-3B. \textbf{Bold} denotes the best and \underline{underline} the second best result per column.}
\label{tab:main}
\small
\setlength{\tabcolsep}{3.5pt}
\begin{tabular}{l ccc cccc c}
\toprule
 \multirow{2}{*}{\textbf{Methods}} & \multicolumn{3}{c}{\textbf{Single-Hop QA}} & \multicolumn{4}{c}{\textbf{Multi-Hop QA}} & \\
\cmidrule(lr){2-4}\cmidrule(lr){5-8}
 ~ & NQ & TriviaQA & PopQA & HotpotQA & 2Wiki & Musique & Bamboogle & Avg. \\
\midrule
\rowcolor{gray!10}
\multicolumn{9}{c}{\textit{w/o Retrieval}} \\
Direct Generation & 0.106 & 0.288 & 0.108 & 0.149 & 0.244 & 0.020 & 0.024 & 0.134 \\
SFT & 0.249 & 0.292 & 0.104 & 0.186 & 0.248 & 0.044 & 0.112 & 0.176 \\
R1-Instruct & 0.210 & 0.449 & 0.171 & 0.208 & 0.275 & 0.060 & 0.192 & 0.224 \\
R1-Base & 0.226 & 0.455 & 0.173 & 0.201 & 0.268 & 0.055 & 0.224 & 0.229 \\
\midrule
\rowcolor{gray!10}
\multicolumn{9}{c}{\textit{w/ Single-Hop Retrieval}} \\
Naive RAG & 0.348 & 0.544 & 0.387 & 0.255 & 0.226 & 0.047 & 0.080 & 0.270 \\
\midrule
\rowcolor{gray!10}
\multicolumn{9}{c}{\textit{w/ Multi-Hop Retrieval}} \\
IRCoT & 0.111 & 0.312 & 0.200 & 0.164 & 0.171 & 0.067 & 0.240 & 0.181 \\
Search-o1 & 0.238 & 0.472 & 0.262 & 0.221 & 0.218 & 0.054 & 0.320 & 0.255 \\
Search-R1-Base & 0.421 & 0.583 & 0.413 & 0.297 & 0.274 & 0.066 & 0.128 & 0.312 \\
Search-R1-Instruct & 0.397 & 0.565 & 0.391 & 0.331 & 0.310 & 0.124 & 0.232 & 0.336 \\
ReSearch-Base & 0.427 & 0.597 & 0.430 & 0.305 & 0.272 & 0.074 & 0.128 & 0.319 \\
ReSearch-Instruct & 0.365 & 0.571 & 0.395 & 0.351 & 0.272 & 0.095 & 0.266 & 0.331 \\
StepSearch-Base & -- & -- & -- & 0.329 & 0.339 & \underline{0.181} & 0.328 & -- \\
StepSearch-Instruct & -- & -- & -- & 0.345 & 0.320 & 0.174 & 0.344 & -- \\
AutoRefine-Base & 0.467 & 0.620 & 0.450 & 0.405 & 0.393 & 0.157 & 0.344 & 0.405 \\
AutoRefine-Instruct & 0.436 & 0.597 & 0.447 & 0.404 & 0.380 & 0.169 & 0.336 & 0.396 \\
MR-Search (Base)& \textbf{0.477} & \underline{0.635} & \underline{0.460} & 0.419 & 0.401 & 0.165 & 0.344 & 0.414 \\
GiGPO (Instruct)  & 0.420 & 0.595 & 0.424 & 0.369 & 0.370 & 0.126 & \textbf{0.641} & \underline{0.421} \\
\rowcolor{cyan!10}
IG-Search-Base & \underline{0.476} & \textbf{0.637} & \textbf{0.476} & \textbf{0.436} & \underline{0.415} & 0.179 & 0.390 & \textbf{0.430} \\
\rowcolor{cyan!10}
IG-Search-Instruct & 0.450 & 0.614 & \underline{0.460} & \underline{0.431} & \textbf{0.440} & \textbf{0.191} & \underline{0.424} & \textbf{0.430} \\
\bottomrule
\end{tabular}
\end{table}
 
\subsection{Ablation Studies}
\label{sec:ablation}

\subsubsection{Component Ablation (RQ2)}
\label{sec:component_ablation}

\par
We systematically remove one component at a time from IG-Search framework. 
All ablations use Qwen2.5-3B-Base. 
Results are presented in Table~\ref{tab:ablation}.

\par
Removing the IG reward entirely reduces the framework to a trajectory-level baseline, decreasing the average EM from 0.430 to 0.403.
\footnote{This single-seed ablation is functionally equivalent to AutoRefine-Base; the 0.2-point gap from Table~\ref{tab:main} is within seed variance (Appendix~\ref{app:statistical}).}
The degradation is most severe on multi-hop benchmarks (Bamboogle: $-$4.9, HotpotQA: $-$3.4), confirming that step-level reward is especially effective when multiple precise search steps are needed.

\par
Replacing asymmetric scaling with symmetric treatment ($\lambda = 1.0$) causes the largest single-component degradation, reducing the average EM to 0.391. 
This may seem surprising, but under $\lambda{=}1.0$, the surviving negative values retain their full magnitude (approximately $[-3, -0.5]$ nats after dead-zone filtering). 
As a result, occasional large negative outliers from early poorly formed queries can dominate individual minibatches. 
This overwhelming penalty drives the policy toward search avoidance, as confirmed by the monotonic drop in search frequency shown in Appendix~\ref{app:lambda_searchfreq}.

\par
Removing the dead zone ($\delta = 0$) reduces performance to 0.416. 
On questions the model can already answer, IG does not faithfully reflect retrieval quality (yielding small positive, near-zero, or even misleading negative penalties; see Case 3 in \S\ref{sec:deadzone_cases})
and feeding these into the advantage interferes with learning on genuinely informative search steps.

\par
Disabling soft clipping yields 0.420 average EM. Although extreme IG values are infrequent, they can dominate the advantage signal in individual minibatches and destabilize gradient updates.

\par
Without query length normalization, the average EM drops to 0.408. The model exploits this by generating abnormally long search queries to accumulate larger total IG bonuses, a form of reward hacking that degrades retrieval effectiveness.

\par
Replacing the default query-only modulation scope with think-and-query modulation reduces the average EM to 0.422, with more pronounced degradation on multi-hop benchmarks (HotpotQA $-$0.8, 2Wiki $-$0.7), where think steps are longer and dilute the IG signal across tokens unrelated to query formulation. Extending modulation to all trajectory tokens further reduces average EM to 0.413, as the IG term contaminates the advantage signal for refine and answer tokens that are already supervised by the trajectory-level reward. These results confirm that IG is effective as a targeted signal applied exclusively to query tokens.

\begin{table}[t]
\centering
\caption{(RQ2) Component ablation on Qwen2.5-3B-Base (Exact Match). The top block removes or modifies individual stabilization components; the bottom block varies the scope of per-token advantage modulation. The first row is the full IG-Search and serves as the reference.}
\label{tab:ablation}
\small
\setlength{\tabcolsep}{3.5pt}
\begin{tabular}{l ccc cccc c}
\toprule
\multirow{2}{*}{\textbf{Configuration}} & \multicolumn{3}{c}{\textbf{Single-Hop QA}} & \multicolumn{4}{c}{\textbf{Multi-Hop QA}} & \multirow{2}{*}{Avg.} \\
\cmidrule(lr){2-4}\cmidrule(lr){5-8}
~ & NQ & TriviaQA & PopQA & HotpotQA & 2Wiki & Musique & Bamboogle & ~ \\
\midrule
IG-Search (full) & \textbf{0.476} & \textbf{0.637} & \textbf{0.476} & \textbf{0.436} & \textbf{0.415} & \textbf{0.179} & \textbf{0.390} & \textbf{0.430} \\
\midrule
w/o IG reward & 0.465 & 0.618 & 0.449 & 0.402 & 0.391 & 0.155 & 0.341 & 0.403 \\
w/o asymmetric scaling & 0.448 & 0.605 & 0.445 & 0.398 & 0.378 & 0.148 & 0.312 & 0.391 \\
w/o dead zone & 0.465 & 0.625 & 0.462 & 0.422 & 0.402 & 0.170 & 0.368 & 0.416 \\
w/o soft clipping & 0.468 & 0.629 & 0.465 & 0.426 & 0.407 & 0.172 & 0.372 & 0.420 \\
w/o query length norm & 0.460 & 0.619 & 0.455 & 0.415 & 0.397 & 0.161 & 0.352 & 0.408 \\
\midrule
w/ think+query modulation & 0.470 & 0.631 & 0.469 & 0.428 & 0.408 & 0.173 & 0.378 & 0.422 \\
w/ all-token modulation & 0.463 & 0.624 & 0.460 & 0.419 & 0.398 & 0.165 & 0.360 & 0.413 \\
\bottomrule
\end{tabular}
\end{table}

\subsubsection{Counterfactual Baseline Design (RQ3)}
\label{sec:baseline_ablation}

A key design choice in IG-Search is the construction of the counterfactual context $\mathcal{C}_t^{\text{rand}}$. We compare four alternatives: 
(1)~\emph{Random docs + refine} (default): both the documents and the refinement are replaced with a randomly sampled pair;
(2)~\emph{Random docs only}: only the documents are replaced while the original refinement is retained; 
(3)~\emph{Bottom-$k$ docs}: the documents are replaced with the least relevant results from the retriever for the same query; 
(4)~\emph{Empty}: the documents and refinement are removed entirely.

\begin{figure}[t]
\centering
\includegraphics[width=0.5\linewidth]{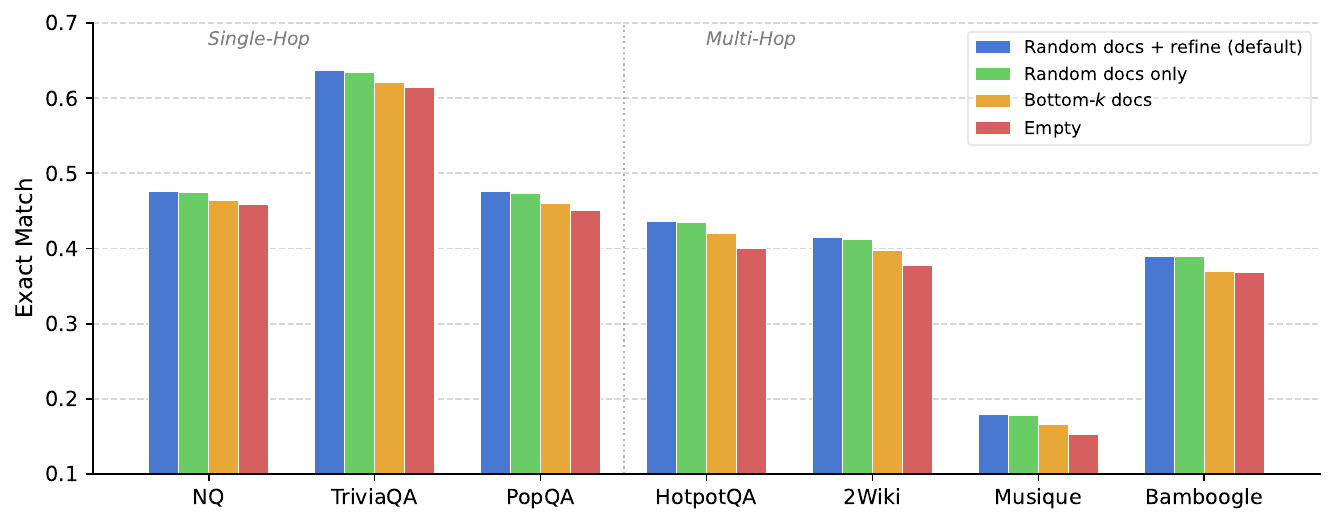}
\caption{(RQ3) Comparison of counterfactual baseline designs across seven benchmarks. The default random docs + refine baseline achieves the highest performance, with the empty baseline showing disproportionately larger degradation on multi-hop benchmarks.}
\label{fig:baseline_design}
\end{figure}

The two random-document variants perform nearly identically (0.430 vs.\ 0.428), indicating that the refinement content in the counterfactual has minimal impact. 
The bottom-$k$ variant achieves 0.414, notably below the random baseline, because bottom-ranked documents from the same query retain some topical overlap, reducing the contrast.
The empty baseline performs worst at 0.403, confirming that length mismatch introduces systematic bias, as discussed in \S\ref{sec:ig}. 
As shown in Figure~\ref{fig:baseline_design}, this degradation is disproportionately concentrated on multi-hop benchmarks, consistent with the hypothesis that length bias accumulates across multiple retrieval steps.

\subsubsection{Hyperparameter Sensitivity (RQ4)}
\label{sec:hyperparam}

\par
\begin{figure}[t]
\centering
\includegraphics[width=\linewidth]{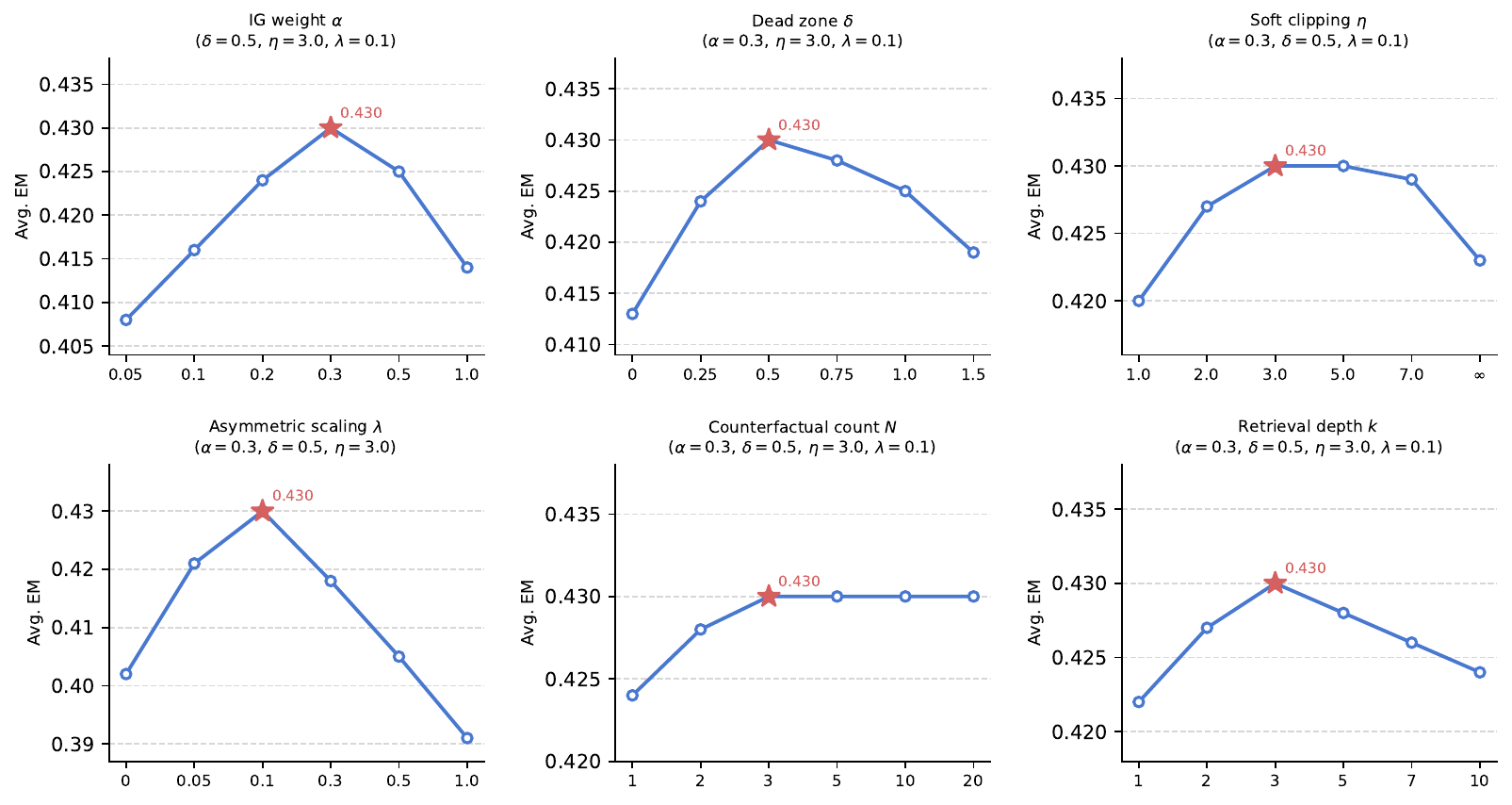}
\caption{(RQ4) Sensitivity of IG-Search to five IG-specific hyperparameters ($\alpha$, $\lambda$, $\delta$, $\eta$, $N$) and to the retrieval depth $k$, on Qwen2.5-3B-Base (average EM across seven benchmarks). Red stars indicate default values.}
\label{fig:hyperparam}
\end{figure}

We analyze the sensitivity of IG-Search to five IG-specific hyperparameters ($\alpha$, $\lambda$, $\delta$, $\eta$, $N$) and to the retrieval depth $k$, varying each while fixing the others at their default values. All experiments use Qwen2.5-3B-Base. Results are shown in Figure~\ref{fig:hyperparam}.

\par
The IG weight $\alpha$ performs best at 0.3; smaller values underweight the step-level signal while larger values cause the IG term to dominate the trajectory-level advantage. 
The asymmetric scaling factor $\lambda$ exhibits the sharpest sensitivity: symmetric treatment ($\lambda{=}1.0$) degrades performance to 0.391, consistent with the learned search-avoidance dynamics tracked in Appendix~\ref{app:lambda_searchfreq}.
The dead zone threshold $\delta$ and soft clipping threshold $\eta$ are comparatively robust across a broad range of values.

\par
For the number of counterfactual contexts $N$, performance improves from $N{=}1$ (0.421) to $N{=}3$ (0.430) and then plateaus, confirming that $N{=}3$ provides a sufficient estimate of the counterfactual baseline without additional overhead. 
For retrieval depth $k$, IG-Search achieves peak performance at $k{=}3$: performance is lower at $k{=}1$ due to insufficient evidence, and degrades gracefully at higher $k$ due to increased noise from less relevant documents.

\subsubsection{Model Size (RQ5)}
\label{sec:model_size}

\par
At the 7B scale, as shown in Table~\ref{tab:model_size}, IG-Search-Instruct achieves the highest average EM of 0.479, surpassing MR-Search (0.460) by 1.9 points and AutoRefine-Base (0.455) by 2.4 points.
GiGPO-7B reaches 0.472, with its advantage concentrated on Bamboogle (0.689 vs.\ 0.560; 125 questions), while IG-Search-Instruct outperforms GiGPO on the remaining six benchmarks.
IG-Search-Base achieves 0.476, also surpassing MR-Search and AutoRefine-Base.
These results demonstrate that IG-Search scales consistently across model sizes.
We extend this analysis to four model sizes in Appendix~\ref{app:scaling}.

\begin{table}[htbp]
\centering
\caption{(RQ5) Exact Match accuracy at the 7B scale (Qwen2.5-7B). \textbf{Bold} denotes the best result per column.}
\label{tab:model_size}
\small
\setlength{\tabcolsep}{3.5pt}
\begin{tabular}{l ccc cccc c}
\toprule
 \multirow{2}{*}{\textbf{Method}} & \multicolumn{3}{c}{\textbf{Single-Hop QA}} & \multicolumn{4}{c}{\textbf{Multi-Hop QA}} & \multirow{2}{*}{Avg.} \\
\cmidrule(lr){2-4}\cmidrule(lr){5-8}
~ & NQ & TriviaQA & PopQA & HotpotQA & 2Wiki & Musique & Bamboogle & ~ \\
\midrule
Search-R1-Base & 0.469 & 0.627 & 0.449 & 0.410 & 0.272 & 0.173 & 0.456 & 0.408 \\
AutoRefine-Base      & 0.484 & 0.659 & 0.487 & 0.451 & 0.405 & 0.187 & 0.512 & 0.455 \\
MR-Search (Base)             & \textbf{0.502} & 0.666 & 0.472 & 0.468 & 0.436 & \textbf{0.221} & 0.452 & 0.460 \\
GiGPO (Instruct)                & 0.464 & 0.647 & 0.461 & 0.416 & 0.436 & 0.189 & \textbf{0.689} & 0.472 \\
\rowcolor{cyan!10} 
IG-Search-Base & 0.501 & \textbf{0.675} & \textbf{0.504} & 0.473 & 0.430 & 0.205 & 0.544 & 0.476 \\
\rowcolor{cyan!10}
IG-Search-Instruct & 0.490 & 0.665 & 0.495 & \textbf{0.480} & \textbf{0.445} & 0.215 & 0.560 & \textbf{0.479} \\
\bottomrule
\end{tabular}
\end{table}

\subsection{Analysis (RQ6)}
\label{sec:analysis}

\subsubsection{Training Dynamics}
\label{sec:training_dynamics}

\par
\begin{figure}[t]
    \centering
    \includegraphics[width=\linewidth]{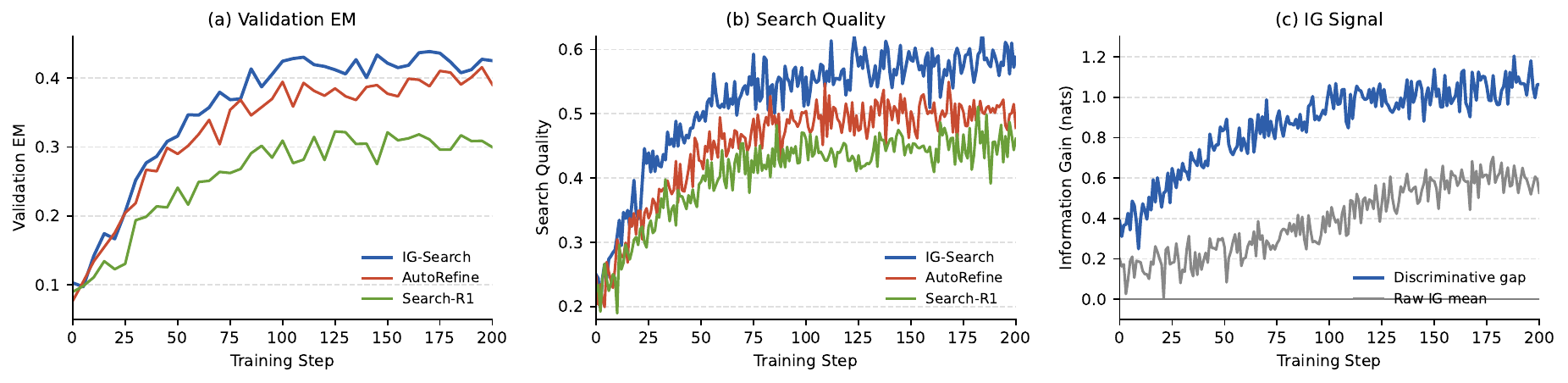}
    \caption{(RQ6) Training dynamics on Qwen2.5-3B-Base. (a)~Validation EM averaged over seven benchmarks. (b)~Search quality, measured as the fraction of retrieval calls whose returned documents contain the gold answer. 
    (c)~Raw per-step IG averaged over all search calls in the batch (a per-position decomposition appears in Figure~\ref{fig:ig_by_step}), and the \emph{discriminative gap} (mean IG on correct minus incorrect trajectories), which stays strictly positive throughout training.}
    \label{fig:training_dynamics}
\end{figure}

Figure~\ref{fig:training_dynamics} tracks training on Qwen2.5-3B-Base, comparing IG-Search against AutoRefine and Search-R1 under identical configurations. 
Panel~(a) shows that IG-Search pulls ahead of both baselines after roughly 40 steps and maintains a widening gap, reaching the final averages reported in Table~\ref{tab:main}. 
Panel~(b) shows that IG-Search consistently produces higher search quality, measured as the fraction of retrieval calls whose documents contain the gold answer, confirming that step-level IG supervision directly improves query formulation. 
Panel~(c) examines the IG signal itself. The raw per-step IG, averaged over all search calls in the batch, rises from roughly $0.15$ nats to $0.61$ nats by step~200. 
Because this average is weighted toward the first hop, we defer the per-position decomposition to Figure~\ref{fig:ig_by_step}.
More importantly, the discriminative gap, defined as the mean IG on eventually-correct trajectories minus that on incorrect ones, stays clearly positive and widens over time, showing that IG reliably separates effective queries from ineffective ones across the full training run.

\subsubsection{IG Reflects Multi-Hop Retrieval Structure}
\label{sec:ig_distribution}

\par
A natural concern with any learned reward is whether it captures something structurally meaningful or merely fits noise correlated with the outcome. 
We address this by decomposing the per-search-step IG by its position within a rollout on Qwen2.5-3B-Base. 
If IG genuinely measures the marginal informational contribution of a retrieval, then the first search, which operates \emph{without any prior retrieved evidence}, should carry more information than later searches, which retrieve on top of already populated evidence.

\begin{figure}[t]
    \centering
    \includegraphics[width=0.5\linewidth]{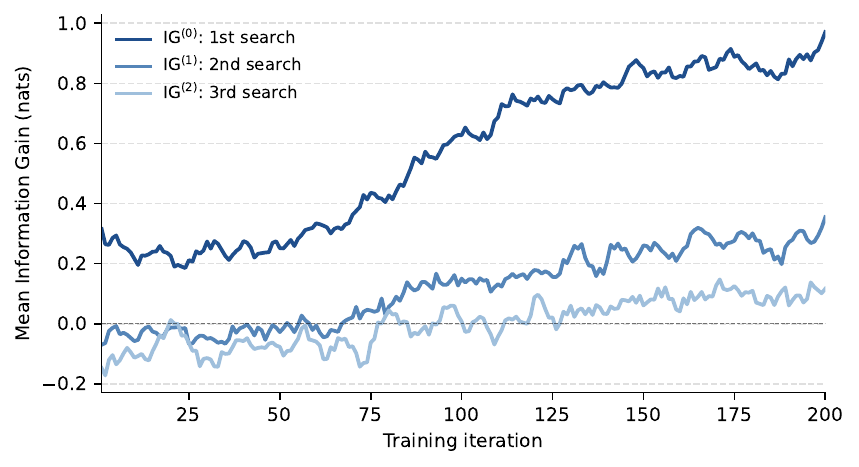}
    \caption{(RQ6) Mean \emph{raw} Information Gain (before dead-zone filtering) decomposed by search-step position $k \in \{0,1,2\}$ during training on Qwen2.5-3B-Base. IG values are reported in nats per answer token. Each curve tracks the average IG of the $(k{+}1)$-th search call across all training questions. The ordering $\mathrm{IG}^{(0)} > \mathrm{IG}^{(1)} > \mathrm{IG}^{(2)}$ holds empirically from the first iteration and the gap persists throughout training, indicating that the signal is aligned with the diminishing-marginal-value structure of multi-hop retrieval.}
    \label{fig:ig_by_step}
\end{figure}

In Figure~\ref{fig:ig_by_step}, we plot the mean IG of the $k$-th search step, $\mathrm{IG}^{(k)}$, for $k \in \{0, 1, 2\}$ throughout training and three key observations stand out. 
First, the ordering $\mathrm{IG}^{(0)} > \mathrm{IG}^{(1)} > \mathrm{IG}^{(2)}$ holds empirically from the first training iteration, before any policy update has shaped the signal, showing that IG is aligned with the diminishing-marginal-value structure of multi-hop retrieval from the outset rather than as a byproduct of optimization.
Second, all three curves rise over training: $\mathrm{IG}^{(0)}$ grows from roughly $0.45$ to $0.92$ nats and $\mathrm{IG}^{(1)}$ from near zero to approximately $0.32$, indicating that IG-Search improves not only the opening query but also the follow-ups, making later searches genuinely additive rather than redundant.
Note that Figure~\ref{fig:ig_by_step} reports raw IG values before dead-zone filtering; because the per-step distribution has high variance, a positional mean below~$\delta$ does not imply that all individual values at that position are filtered.
Third, the gap between positions does not collapse. A reward-hacking policy that inflated IG uniformly across all positions would produce converging curves; the persistent separation instead shows that the policy continues to respect the structural asymmetry between hops. 
Together these dynamics indicate that the signal IG-Search optimizes is aligned with the information flow of multi-hop reasoning rather than being an arbitrary scalar that happens to correlate with answer correctness.

\subsubsection{Dead Zone Case Study}
\label{sec:deadzone_cases}
 
Figure~\ref{fig:deadzone_cases} presents three cases from training trajectories illustrating the necessity of dead zone filtering.
 
Case~1 involves a question the model can already answer. Although the retrieved documents are relevant, they provide no new information, resulting in a small positive IG (0.13). Without the dead zone, this weak signal would reinforce the query despite its quality being irrelevant to the outcome.
 
Case~3 illustrates a more subtle failure mode driven by parametric knowledge. The real documents mention several years (1887, 1889, the 1889 World's Fair), diluting the model's confidence in ``1889'' specifically, while random documents leave the parametric prior intact and yield a higher probability on the gold token. The result is a \emph{negative} IG ($-$0.24) despite the query being well-formed. Without the dead zone, this correct and relevant query would be erroneously penalized.

In contrast, Case~2 shows a genuine knowledge gap where the model cannot answer without retrieval, and the high IG (1.72) accurately reflects the query's effectiveness. 
Together these cases show that on model-proficient questions IG does not faithfully measure retrieval quality, and the dead zone is essential for filtering out such spurious signals.

\begin{figure}[t]
\centering
\small
\setlength{\tabcolsep}{3pt}
\begin{tabular}{p{0.96\textwidth}}
\toprule
\textbf{Case 1: Model-proficient question, small positive IG (filtered)} \\
\midrule
\textbf{Q:} What is the capital of France? \quad \textbf{Gold:} Paris \\[3pt]
\texttt{<think>} The capital of France is Paris. Let me verify. \texttt{</think>} \\
\texttt{<search>} capital of France \texttt{</search>} \\
\texttt{<documents>} Paris is the capital and most populous city of France\ldots \texttt{</documents>} \\
\texttt{<refine>} The capital of France is Paris. \texttt{</refine>} \quad \texttt{<answer>} Paris \texttt{</answer>} \\[4pt]
\colorbox{gray!10}{\parbox{\dimexpr\linewidth-2\fboxsep\relax}{\strut \textbf{\textit{IG computation:}} \\
$\log \pi_{\theta}(\text{Paris} \mid \mathcal{C}^{\text{real}}) = -0.15$, \quad $\frac{1}{N} \sum_j \log \pi_{\theta}(\text{Paris} \mid \mathcal{C}^{\text{rand, j}}) = -0.28$ \quad $\Rightarrow$ \textbf{IG = 0.13}\strut}} \\[2pt]
\colorbox{green!12}{\parbox{\dimexpr\linewidth-2\fboxsep\relax}{\strut Dead zone ($\delta{=}0.5$): \textbf{filtered}. small positive IG from a model-proficient question; no spurious reinforcement.\strut}} \\
\midrule
\textbf{Case 2: Knowledge-gap question, high IG (passed)} \\
\midrule
\textbf{Q:} Who is the father of the father of modern experimental psychology? \quad \textbf{Gold:} Maximilian Wundt \\[3pt]
\texttt{<think>} I need to find the father of modern experimental psychology first. \texttt{</think>} \\
\texttt{<search>} father of modern experimental psychology \texttt{</search>} \\
\texttt{<documents>} Wilhelm Wundt is widely regarded as the father of experimental psychology. He was born on 16 August 1832 to parents Maximilian Wundt\ldots \texttt{</documents>} \\
\texttt{<refine>} Wilhelm Wundt is the father of experimental psychology. His father was Maximilian Wundt. \texttt{</refine>} \quad \texttt{<answer>} Maximilian Wundt \texttt{</answer>} \\[4pt]
\colorbox{gray!10}{\parbox{\dimexpr\linewidth-2\fboxsep\relax}{\strut \textbf{\textit{IG computation:}} \\
$\log \pi_{\theta}(\text{M.\ Wundt} \mid \mathcal{C}^{\text{real}}) = -1.82$, \quad $\frac{1}{N} \sum_j \log \pi_{\theta}(\text{M.\ Wundt} \mid \mathcal{C}^{\text{rand, j}}) = -3.54$ \quad $\Rightarrow$ \textbf{IG = 1.72}\strut}} \\[2pt]
\colorbox{blue!8}{\parbox{\dimexpr\linewidth-2\fboxsep\relax}{\strut Dead zone ($\delta{=}0.5$): \textbf{passed}. Genuine knowledge gap; high IG correctly rewards this query.\strut}} \\
\midrule
\textbf{Case 3: Model-proficient question, spurious negative IG (filtered)} \\
\midrule
\textbf{Q:} When was the Eiffel Tower completed? \quad \textbf{Gold:} 1889 \\[3pt]
\texttt{<search>} Eiffel Tower completion year \texttt{</search>} \\
\texttt{<documents>} The Eiffel Tower was constructed from 1887 to 1889 as the centerpiece of the 1889 World's Fair\ldots \texttt{</documents>} \\
\texttt{<refine>} The Eiffel Tower was completed in 1889. \texttt{</refine>} \quad \texttt{<answer>} 1889 \texttt{</answer>} \\[4pt]
\colorbox{gray!10}{\parbox{\dimexpr\linewidth-2\fboxsep\relax}{\strut \textbf{\textit{IG computation:}} \\
$\log \pi_{\theta}(\text{1889} \mid \mathcal{C}^{\text{real}}) = -0.42$, \quad $\frac{1}{N} \sum_j \log \pi_{\theta}(\text{1889} \mid \mathcal{C}^{\text{rand, j}}) = -0.18$ \quad $\Rightarrow$ \textbf{IG = $-$0.24}\strut}} \\[2pt]
\colorbox{red!8}{\parbox{\dimexpr\linewidth-2\fboxsep\relax}{\strut Without dead zone: this correct, relevant query would be \emph{penalized} because real documents introduce competing facts that dilute the parametric prior.\strut}} \\
\colorbox{green!12}{\parbox{\dimexpr\linewidth-2\fboxsep\relax}{\strut Dead zone ($\delta{=}0.5$): \textbf{filtered}. Spurious negative signal suppressed; no incorrect penalization.\strut}} \\
\bottomrule
\end{tabular}
\captionof{figure}{Dead zone filtering on three illustrative training cases. Cases 1 and 3 are model-proficient questions where raw IG fails to reflect retrieval quality and are both suppressed by the dead zone; Case 2 is a genuine knowledge gap where high IG correctly rewards the query and passes the filter.}
\label{fig:deadzone_cases}
\end{figure}

\subsubsection{Search Behavior Across Question Types}
\label{sec:search_behavior}

\par We examine whether IG-Search learns to calibrate its search depth to the complexity of the question. On single-hop benchmarks, the trained policy issues on average 1.3 search calls per rollout, whereas on multi-hop benchmarks it issues 2.4. 
By contrast, the trajectory-level baseline AutoRefine-Base uses 2.0 and 2.3 search calls respectively on the same two categories. IG-Search thus exhibits a more task-adaptive search pattern: it is much more conservative on questions that can be resolved by a single retrieval, while remaining active on questions that genuinely require multi-hop evidence.

\par
\begin{figure}[t]
    \centering
    \includegraphics[width=\linewidth]{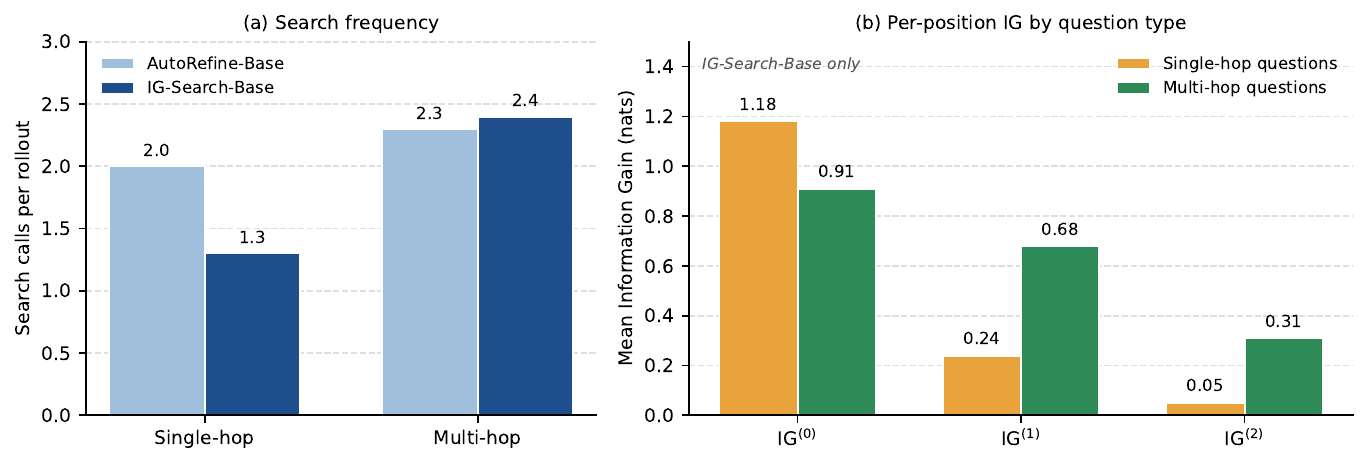}
    \caption{(RQ6) Search behavior of IG-Search-Base across question types. (a)~Average number of search calls per rollout on single-hop vs.\ multi-hop benchmarks, compared with AutoRefine-Base. (b)~Mean Information Gain at each search-step position, split by question type. On single-hop questions, nearly all information is captured by the first search; on multi-hop questions, subsequent searches continue to contribute.}
    \label{fig:search_behavior}
\end{figure}

 
We compute per-position IG on the evaluation benchmarks
at the final checkpoint. Figure~\ref{fig:search_behavior} further breaks down the per-position IG on the two categories. On single-hop questions nearly all of the information gain is concentrated in the opening search ($\mathrm{IG}^{(0)} = 1.18$), while any subsequent retrieval contributes less ($\mathrm{IG}^{(1)} = 0.24$), indicating that a second search is usually an unnecessary verification. 
On multi-hop questions the signal is distributed more evenly across steps ($\mathrm{IG}^{(0)} = 0.91$, $\mathrm{IG}^{(1)} = 0.68$, $\mathrm{IG}^{(2)} = 0.31$), showing that each additional hop continues to bring in new evidence. 
These observations jointly indicate that IG-Search does not merely learn to search more; it learns \emph{when} to search and when to stop.
 
\subsubsection{Learning Signal in the All-Failure Scenario}
\label{sec:all_failure}
\par A structural advantage of IG-Search is its ability to provide training signal even when every sampled trajectory in a group produces an incorrect answer, a scenario common in early training and on harder multi-hop questions. 
When all rollouts share the same outcome reward, the group-normalized advantages $\hat{A}_i$ collapse toward zero, causing trajectory-level rewards to fail to provide meaningful learning signals. 
IG-Search bypasses this failure mode because the per-token IG modulation $\alpha \cdot \widetilde{\mathrm{IG}}_t / |\mathcal{Q}_t|$ is added independently of answer correctness: tokens from an informative search query receive positive modulation, while those from an uninformative query receive either zero (filtered by the dead zone) or a small negative signal.
This ensures that the model consistently assigns higher credit to the more effective query.

\par 
Figure~\ref{fig:all_failure_case} illustrates this on an all-failure batch where two rollouts end in different wrong answers but arrive there through searches of very different quality. 
Rollout A issues a precise opening query that retrieves the answer-bearing document ($\mathrm{IG}^{(0)}_A = 1.58$) and fails at subsequent reasoning, while Rollout B issues a vague query whose retrieved documents are only tangentially related ($\mathrm{IG}^{(0)}_B = 0.08$). 
Trajectory-level rewards assign them zero advantages, but IG-Search continues to differentiate their query tokens and push the policy toward the more informative formulation.

\par
Quantitatively, all-failure questions (every rollout yields $r_{\text{F1}}{=}0$) account for approximately 58\% of the training batch at step~1 and still 10\% at step~200; on this subset the mean absolute IG modulation $\alpha \cdot |\widetilde{\mathrm{IG}}_t|$ remains around 0.21 throughout training, providing a consistent per-token gradient where $\hat{A}_i$ is identically zero.

\begin{figure}[t]
\centering
\small
\setlength{\tabcolsep}{3pt}
\begin{tabular}{p{0.96\textwidth}}
\toprule
\textbf{Question:} Who directed the film that won the Academy Award for Best Picture in 1994? \\
\textbf{Gold Answer:} \textcolor{goldanswer}{\textbf{Robert Zemeckis}} \\
\textit{Batch status: all 5 rollouts answered incorrectly.} \\
\midrule
\textbf{Rollout A} (informative search, wrong final answer) \\[3pt]
\texttt{<think>} I need to find which film won Best Picture at the 1994 Academy Awards, then its director. \texttt{</think>} \\
\texttt{<search>} 1994 Academy Award Best Picture winner \texttt{</search>} \\
\texttt{<documents>} \textcolor{goldanswer}{Forrest Gump}, directed by Robert Zemeckis, won the Academy Award for Best Picture at the 67th Academy Awards held in March 1995, honoring films of 1994\ldots \texttt{</documents>} \\
\texttt{<refine>} Forrest Gump won Best Picture for 1994. \texttt{</refine>} \\
\texttt{<search>} Forrest Gump director \texttt{</search>} \\
\texttt{<documents>} Forrest Gump is a 1994 American comedy-drama film based on the 1986 novel by Winston Groom\ldots The film was produced by Wendy Finerman\ldots \texttt{</documents>} \\
\texttt{<refine>} Forrest Gump was produced by Wendy Finerman. \texttt{</refine>} \quad \texttt{<answer>} Wendy Finerman \texttt{</answer>} \\[4pt]
\colorbox{gray!10}{\parbox{\dimexpr\linewidth-2\fboxsep\relax}{\strut \textbf{\textit{IG computation:}} $\mathrm{IG}^{(0)}_A = 1.58$ (retrieved document explicitly mentions the gold answer), $\mathrm{IG}^{(1)}_A = 0.12$ (second retrieval loses focus). \\ \textbf{\textit{Trajectory reward:}} $r_{\text{F1}} = 0$. \textit{GRPO advantage:} $\hat{A}_A \approx 0$.\strut}} \\[2pt]
\colorbox{blue!8}{\parbox{\dimexpr\linewidth-2\fboxsep\relax}{\strut IG-Search: opening query tokens receive positive modulation from $\mathrm{IG}^{(0)}_A$, reinforcing the precise query formulation.\strut}} \\
\midrule
\textbf{Rollout B} (uninformative search, wrong final answer) \\[3pt]
\texttt{<think>} Let me look up Best Picture winners. \texttt{</think>} \\
\texttt{<search>} best picture movies \texttt{</search>} \\
\texttt{<documents>} The Academy Award for Best Picture is one of the Academy Awards presented annually by the Academy of Motion Picture Arts and Sciences\ldots The first winner was Wings in 1929\ldots \texttt{</documents>} \\
\texttt{<refine>} Best Picture has been awarded since 1929. \texttt{</refine>} \\
\texttt{<search>} movie directors 1990s \texttt{</search>} \\
\texttt{<documents>} Notable directors of the 1990s include Quentin Tarantino, Martin Scorsese, Steven Spielberg\ldots \texttt{</documents>} \\
\texttt{<refine>} Steven Spielberg is a notable 1990s director. \texttt{</refine>} \quad \texttt{<answer>} Steven Spielberg \texttt{</answer>} \\[4pt]
\colorbox{gray!10}{\parbox{\dimexpr\linewidth-2\fboxsep\relax}{\strut \textbf{\textit{IG computation:}} $\mathrm{IG}^{(0)}_B = 0.08$ (generic information, no year-specific content), $\mathrm{IG}^{(1)}_B = -0.04$ (retrieval drifts further from gold). \\ \textbf{\textit{Trajectory reward:}} $r_{\text{F1}} = 0$. \textit{GRPO advantage:} $\hat{A}_B \approx 0$.\strut}} \\[2pt]
\colorbox{red!8}{\parbox{\dimexpr\linewidth-2\fboxsep\relax}{\strut IG-Search: opening query tokens receive near-zero modulation, offering no reinforcement to the vague formulation.\strut}} \\
\bottomrule
\end{tabular}
\captionof{figure}{(RQ6) Case study from an all-failure batch (all five rollouts receive $r_{\text{F1}} = 0$ and thus $\hat{A}_i \approx 0$). Rollouts A and B receive identical trajectory-level advantages despite searches of very different quality; IG-Search differentiates them via per-step IG, providing nonzero gradient signal for the query tokens and pushing the policy toward the more informative formulation.}
\label{fig:all_failure_case}
\end{figure}

\section{Limitations}
\label{sec:limitations}

\par Despite the strong performance of IG-Search, several limitations remain for future investigation. First, the IG computation requires access to the gold answer during training to evaluate $\log \pi_\theta(a^* \mid \mathcal{C})$, which restricts applicability to settings where ground-truth supervision is available. Developing answer-free proxies for information gain, such as substituting the majority-vote answer across the rollout group in place of the gold answer, is a promising direction for weakly supervised settings.

\par Second, the stabilization hyperparameters ($\delta$, $\eta$, $\lambda$, $\alpha$) were tuned on NQ and HotpotQA and held fixed across all experiments. Although the sensitivity analysis in \S\ref{sec:hyperparam} shows that IG-Search is robust within reasonable ranges, we have not investigated whether task families with very different answer structures, such as numerical reasoning, temporal reasoning, or long-form answers, would benefit from a different parameter tuning strategy

\par Finally, the retrieval component uses an offline Wikipedia corpus and therefore lacks current or time-sensitive information. This is a shared limitation of the retrieval-augmented reasoning literature rather than one specific to IG-Search, but it does limit the real-world applicability of models trained under our setup.

\section{Conclusion and Future Work}
\label{sec:conclusion}

\par 
We have presented IG-Search, a reinforcement learning framework that assigns a distinct reward to each individual search step. 
This reward measures how much the retrieved documents improve the policy's confidence in the gold answer relative to a random counterfactual baseline.
This reformulation turns a sparse, trajectory-level signal into a dense, query-grounded one, enabling per-token credit assignment across individual search steps within a rollout. 
Experiments on seven QA benchmarks show that IG-Search achieves the strongest results among open-source models of comparable size at both the 3B and 7B scales, with larger gains on multi-hop benchmarks where trajectory-level credit assignment is more brittle.

\par Future work will explore answer-free variants of the IG signal for weakly supervised settings, extend IG-Search to dynamic retrieval environments such as live web search, and investigate combinations with cross-episode exploration strategies like self-reflection~\citep{xiao2025mrsearch}, which operate at a complementary level of granularity.
\bibliographystyle{plainnat}
\bibliography{reference}

\newpage
\appendix

\section{Related Work}
\label{sec:related_work}

\par
IG-Search sits at the intersection of RL-based reasoning, retrieval-augmented generation, and step-level reward design. 
We review each branch of work below, focusing on how existing methods handle credit assignment for retrieval actions.

\par
\subsection{RL for LLM Reasoning and Retrieval-Augmented Generation}
 Reinforcement learning has become a central tool for improving reasoning in LLMs~\citep{qwen2025,deepseek2025r1,openai2024openaio1card}. Group-relative methods such as GRPO~\citep{shao2024deepseekmath,deepseek2025r1} drive substantial gains on mathematical reasoning and code generation using only outcome-based rewards, and IG-Search builds directly on this framework.

\par
A parallel line of work equips LLMs with retrieval tools to access external knowledge during reasoning~\citep{lewis2020naiverag,jiang2023activerag,ram2023incontext}. 
Early methods use supervised fine-tuning to teach models when and how to search~\citep{asai2024selfrag,shi2023replug,yan2024corrective}, while recent work adopts RL to enable a \emph{search-during-think} paradigm in which the model autonomously invokes retrieval within its reasoning trajectories~\citep{jin2025searchr1,chen2025research,song2025r1searcher,li2025searcho1}. 
AutoRefine~\citep{shi2025autorefine} adds an explicit refinement step and a retrieval-specific reward, and MR-Search~\citep{xiao2025mrsearch} introduces cross-episode self-reflection.

\subsection{Step-Level Reward Design for Retrieval-Augmented Reasoning}
Recent work addresses the limitations of trajectory-level credit assignment along several complementary axes, which we organize by the source of their step-level signal.

\paragraph{Temporal-difference approaches.}
IGPO~\citep{wang2026igpo} defines a turn-level reward as the temporal difference in the policy's gold-answer log-probability between consecutive turns, and uses the resulting turn-level returns in place of the GRPO advantage.
IG-Search shares the intuition that the rollout policy's confidence in the gold answer is a useful source of step-level supervision, but realizes it through a counterfactual comparison at each retrieval step, which isolates the contribution of the retrieved documents and allows the signal to be routed back specifically to the query tokens that produced them.
Because the two methods evaluate under different retrieval environments and metrics, we provide a cross-protocol comparison in Appendix~\ref{app:igpo_comparison}.

\paragraph{Annotation-dependent and state-recurrence methods.}
StepSearch~\citep{li2025stepsearch} computes a TF-IDF coverage reward over golden supporting documents and additionally uses GPT-4o-generated sub-question keywords, requiring intermediate annotations that IG-Search avoids entirely. 
GiGPO~\citep{feng2025gigpo} takes an annotation-free approach by retroactively grouping actions from identical environment states across trajectories and computing a relative advantage within each group. 
Its signal density, however, depends on how often rollouts revisit the same state. 
IG-Search makes the opposite trade-off: it uses the gold answer for a denser, query-grounded signal at every search step regardless of state recurrence. 
Combining these two is a promising direction for future work.
 
\paragraph{Tree-structured credit assignment.}
Tree-GRPO~\citep{ding2025treegrpo} replaces chain-based rollouts with agent-step-level tree search, deriving implicit step-level preference signals from the tree structure itself. 
Because it modifies the sampling strategy rather than the reward signal, it is orthogonal to IG-Search and the two could in principle be combined.

\paragraph{Teacher-dependent pipelines.}
Thinker~\citep{xu2025thinker} and SmartSearch~\citep{wen2026smartsearch} achieve strong results but rely on large teacher models (72B and 32B respectively) and multi-stage training pipelines (SFT$\to$DPO$\to$GRPO or distilled process reward models), whereas IG-Search operates in a single GRPO loop using only standard question-answer pairs.

\paragraph{Relation to process reward models.}
More broadly, IG-Search can be viewed as a retrieval-grounded process reward~\citep{lightman2023letsverifystepstep,zhang2025lessonsdevelopingprocessreward}. 
Process reward models have shown substantial gains on mathematical reasoning by providing step-level supervision, but they typically require a separately trained reward model and, in some cases, human-annotated step labels. 
IG-Search obtains a step-level signal without relying on any of these external components. 
The gold answer and the policy's own log-probabilities are sufficient to generate a reward that is just as detailed as a learned process reward. 
Furthermore, this approach has the advantage of evaluating actual retrieval outcomes directly, rather than relying on an external judge's assessment.

\section{IG-Search Training Algorithm}
\label{app:training_algorithm}

We here provide the complete step-by-step training pipeline for IG-Search framework. 
Algorithm~\ref{alg:ig_search} details how the standard trajectory-level rewards are integrated with our proposed step-level Information Gain computation and advantage modulation within the GRPO loop.

\begin{algorithm}[h]
\footnotesize
\caption{IG-Search Training}
\label{alg:ig_search}
\begin{algorithmic}[1]
\REQUIRE Training set $\mathcal{D}$, policy $\pi_\theta$, search engine $\mathcal{E}$, group size $G$, IG weight $\alpha$
\FOR{each training iteration}
    \FOR{each question $(q, a^*)$ in minibatch}
        \STATE Sample $G$ trajectories $\{o_1, \ldots, o_G\}$ from $\pi_{\theta_{\text{old}}}(\cdot \mid q)$
        \STATE Compute trajectory rewards $R_i = r_{\text{F1}}(o_i) + r_{\text{ret}}(o_i)$
        \STATE Compute advantages $\hat{A}_i$ via Eq.~\ref{eq:grpo_adv}
        \FOR{each trajectory $o_i$}
            \FOR{each search step $t$ in $o_i$}
                \STATE Construct $\mathcal{C}_t^{\text{real}}$ and $\{\mathcal{C}_t^{\text{rand},j}\}_{j=1}^N$
                \STATE Compute $\text{IG}_t$ via Eq.~\ref{eq:ig}
                \STATE Apply stabilization: dead zone, asymmetric scaling, soft clipping $\to$ $\widetilde{\text{IG}}_t$
            \ENDFOR
            \STATE Modulate token-level advantages via Eq.~\ref{eq:adv_mod}
        \ENDFOR
    \ENDFOR
    \STATE Update $\pi_\theta$ using GRPO with modulated advantages $\{\tilde{A}_{i,p}\}$
\ENDFOR
\end{algorithmic}
\end{algorithm}

\section{Statistical Significance}
\label{app:statistical}

To validate the stability of our results, we train IG-Search-Base and AutoRefine-Base with three random seeds on Qwen2.5-3B-Base. The Exact Match scores reported in Table~\ref{tab:main} for these two methods are averaged over the three seeds; Table~\ref{tab:seeds} additionally reports the standard deviations. A paired $t$-test over the seven benchmarks yields $p < 0.01$ for IG-Search-Base versus AutoRefine-Base, and the error bars are non-overlapping on all four multi-hop benchmarks, confirming that the improvements are statistically significant and not an artifact of seed variance.

\begin{table}[h]
\centering
\caption{Exact Match (mean$_{\text{std}}$) across three random seeds on Qwen2.5-3B-Base. Mean values correspond to those reported in Table~\ref{tab:main}.}
\label{tab:seeds}
\small
\setlength{\tabcolsep}{3pt}
\begin{tabular}{l ccc cccc c}
\toprule
 \multirow{2}{*}{\textbf{Method}} & \multicolumn{3}{c}{\textbf{Single-Hop QA}} & \multicolumn{4}{c}{\textbf{Multi-Hop QA}} & \multirow{2}{*}{Avg.} \\
\cmidrule(lr){2-4}\cmidrule(lr){5-8}
~ & NQ & TriviaQA & PopQA & HotpotQA & 2Wiki & Musique & Bamboogle & ~ \\
\midrule
AutoRefine-Base
  & $0.467_{.008}$ & $0.620_{.006}$ & $0.450_{.009}$
  & $0.405_{.010}$ & $0.393_{.011}$ & $0.157_{.007}$ & $0.344_{.015}$
  & $0.405_{.008}$ \\
IG-Search-Base
  & $0.476_{.006}$ & $0.637_{.005}$ & $0.476_{.007}$
  & $0.436_{.008}$ & $0.415_{.009}$ & $0.179_{.005}$ & $0.390_{.011}$
  & $0.430_{.005}$ \\
\bottomrule
\end{tabular}
\end{table}

\section{Search Avoidance under Symmetric Negative Scaling}
\label{app:lambda_searchfreq}

The $\lambda$ ablation in Table~\ref{tab:ablation} and the sensitivity curve in Figure~\ref{fig:hyperparam} both show that setting $\lambda = 1.0$ causes a substantially larger performance drop than any other single-component ablation. We attribute this degradation to \emph{search avoidance}: the policy learns to stop issuing search queries altogether, thereby escaping the negative IG signal. To verify this, we track the average number of search calls per rollout throughout training under the two settings, with all other hyperparameters held at their defaults.

\begin{figure}[h]
\centering
\includegraphics[width=0.5\linewidth]{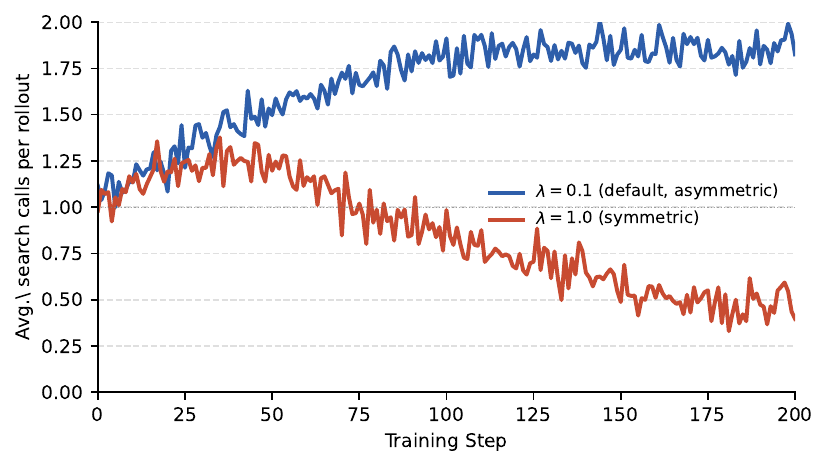}
\caption{Average number of search calls per rollout during training on Qwen2.5-3B-Base, comparing the default asymmetric setting ($\lambda = 0.1$) against symmetric negative scaling ($\lambda = 1.0$). The two variants share the same initialization and exhibit similar exploration in the first $\sim$30 steps; the symmetric variant then decays monotonically toward 1.0 as full-strength negative IG penalties suppress search usage, while the default setting stabilizes around 1.85.}
\label{fig:lambda_search_freq}
\end{figure}

Figure~\ref{fig:lambda_search_freq} reports the result on Qwen2.5-3B-Base. Both runs start from the same initialization and exhibit similar exploration in the first $\sim$30 steps, after which they diverge sharply: the default run stabilizes at approximately 1.85 search calls per rollout, while the symmetric run decays monotonically toward 1.0, as full-strength negative IG on early poorly-formed queries progressively suppresses the policy's tendency to search. This provides direct mechanistic evidence that the performance drop under $\lambda=1.0$ is mediated by learned search avoidance, justifying the asymmetric design.

\section{Scaling Across Model Sizes}
\label{app:scaling}
\par
\begin{figure}[t]
\centering
\includegraphics[width=0.5\linewidth]{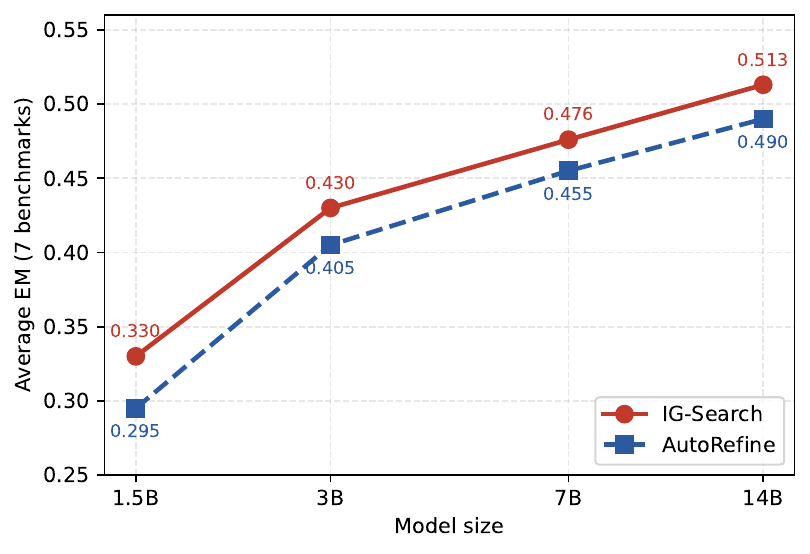}
\caption{Average EM at step 200 across four Qwen2.5 model sizes. IG-Search yields consistent gains over AutoRefine at every scale, demonstrating that step-level IG supervision scales effectively with model capacity.}
\label{fig:scaling}
\end{figure}

We train both methods on Qwen2.5 at four scales under identical settings and report average EM across the seven benchmarks at step 200 (Figure~\ref{fig:scaling}). 
IG-Search delivers consistent gains over AutoRefine at every scale, with improvements of +3.5, +2.5, +2.1, and +2.3 points at 1.5B, 3B, 7B, and 14B respectively. 
The persistent advantage at larger scales indicates that dense step-level IG supervision remains beneficial as model capacity grows, rather than being a low-capacity crutch that stronger policies can recover on their own. 
This confirms that IG-Search scales effectively with model size and provides complementary signal beyond what trajectory-level rewards offer at any scale.

\section{Dataset Statistics}
\label{app:datasets}

Table~\ref{tab:datasets} summarizes the seven question answering benchmarks used in our experiments. All datasets are sourced from the FlashRAG datasets collection. We construct the training set by combining the train splits of NQ and HotpotQA, following~\cite{jin2025searchr1}. For evaluation we use the test split where available (NQ, TriviaQA, PopQA, Bamboogle) and the dev split otherwise (HotpotQA, 2Wiki, Musique).

\begin{table}[h]
\centering
\caption{Statistics of the seven benchmarks used in this paper.}
\label{tab:datasets}
\small
\setlength{\tabcolsep}{5pt}
\begin{tabular}{l ccc c}
\toprule
\multirow{2}{*}{\textbf{Dataset}} & \multicolumn{3}{c}{\textbf{Split sizes}} & \multirow{2}{*}{\textbf{Eval split}} \\
\cmidrule(lr){2-4}
~ & Train & Dev & Test & ~ \\
\midrule
NQ         & 79{,}168 &  8{,}757 &  3{,}610 & test \\
TriviaQA   & 78{,}785 &  8{,}837 & 11{,}313 & test \\
PopQA      & --       & --       & 14{,}267 & test \\
HotpotQA   & 90{,}447 &  7{,}405 & --       & dev  \\
2Wiki      & 15{,}000 & 12{,}576 & --       & dev  \\
Musique    & 19{,}938 &  2{,}417 & --       & dev  \\
Bamboogle  & --       & --       &      125 & test \\
\bottomrule
\end{tabular}
\end{table}

\section{Hyperparameters}
\label{app:hparams}

Table~\ref{tab:hparams} lists the complete set of hyperparameters used in our experiments. The IG-specific hyperparameters ($\alpha, \delta, \lambda, \eta, N$) were selected on NQ and HotpotQA; their sensitivity is analyzed in §\ref{sec:hyperparam}. All other values follow the default configuration of the veRL framework~\citep{Sheng2025hybridflow}.

\begin{table}[h]
\centering
\caption{Complete hyperparameter configuration for IG-Search training.}
\label{tab:hparams}
\small
\begin{tabular}{l l}
\toprule
\textbf{Hyperparameter} & \textbf{Value} \\
\midrule
\rowcolor{gray!10}
\multicolumn{2}{c}{\textit{\textbf{Training configuration}}} \\
Group size $G$                     & 5 \\
Training batch size                & 512 \\
Actor learning rate                & $1 \times 10^{-6}$ \\
Total training steps               & 200 \\
KL coefficient $\beta$             & 0.001 \\
Clip ratio $\epsilon$              & 0.2 \\
Max search calls per rollout       & 5 \\
Rollout temperature                & 1.0 \\
\midrule
\rowcolor{gray!10}
\multicolumn{2}{c}{\textit{\textbf{Retrieval configuration}}} \\
Retriever                          & E5-base-v2 \\
Corpus                             & Wikipedia (Dec.~2018) \\
Top-$k$ documents per query        & 3 \\
\midrule
\rowcolor{gray!10}
\multicolumn{2}{c}{\textit{\textbf{IG-Search specific}}} \\
IG weight $\alpha$                 & 0.3 \\
Dead zone threshold $\delta$       & 0.5 \\
Asymmetric negative scale $\lambda$ & 0.1 \\
Soft clipping threshold $\eta$     & 3.0 \\
Number of counterfactual contexts $N$ & 3 \\
Max context length for IG         & 8192 \\
Max gold answer variants          & 3 \\
\bottomrule
\end{tabular}
\end{table}

\section{Training Overhead}
\label{app:overhead}

Table~\ref{tab:overhead} reports the per-step wall-clock time breakdown. The IG computation adds approximately 6.4\% to total step time. 
\textbf{This overhead is training-only and does not affect inference latency.}

\begin{table}[h]
\centering
\caption{Per-step wall-clock time on 8$\times$H800 with Qwen2.5-3B-Base (seconds, averaged over steps 2--5).}
\label{tab:overhead}
\small
\begin{tabular}{l cc}
\toprule
\textbf{Component} & \textbf{AutoRefine} & \textbf{IG-Search} \\
\midrule
Rollout generation       & 182 & 182 \\
Reference log-prob       &   9 &   9 \\
Reward + KL + Advantage  &   7 &   7 \\
IG computation           &  -- &  15 \\
Policy update            &  38 &  38 \\
\midrule
Total per step           & 236 & 251 \\
IG overhead              &  -- & +6.4\% \\
\bottomrule
\end{tabular}
\end{table}

\section{Raw IG Distribution}
\label{app:ig_dist}

\par
\begin{figure}[h]
\centering
\includegraphics[width=\linewidth]{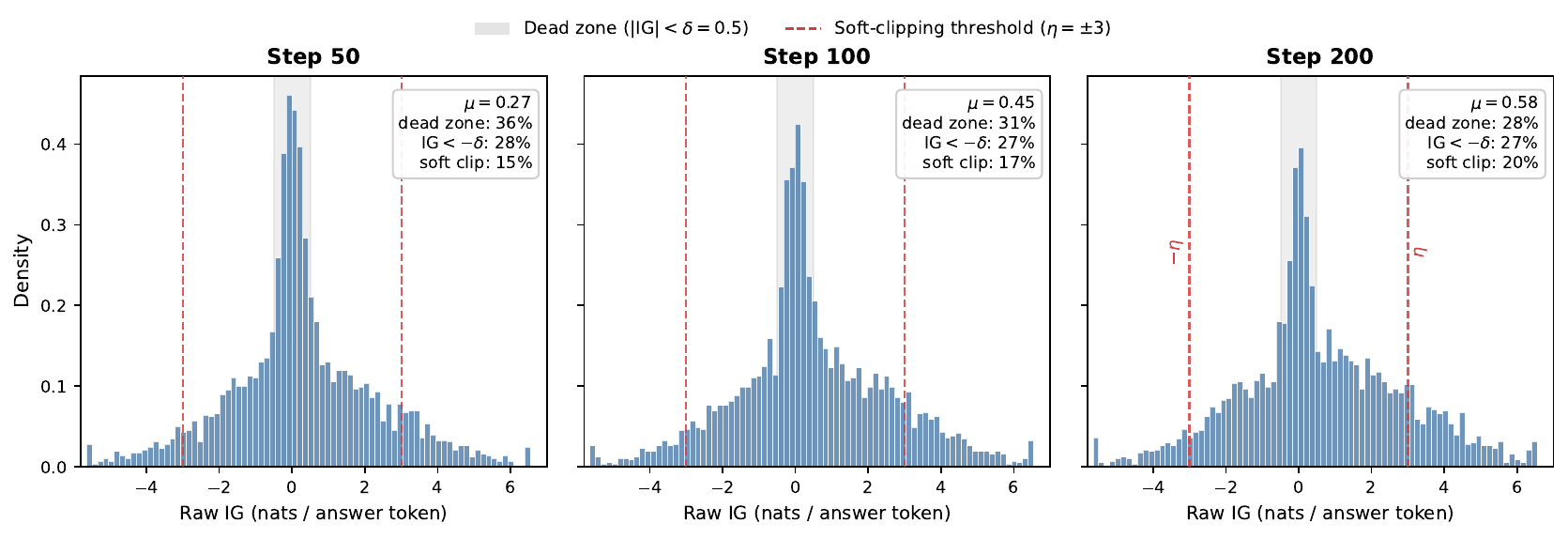}
\caption{Raw IG distribution at three training checkpoints on Qwen2.5-3B-Base.  
The gray band marks the dead-zone region ($|\mathrm{IG}|<\delta{=}0.5$); red dashed lines mark the soft-clipping threshold ($\eta{=}\pm 3$).  
Statistics boxes report the batch mean, dead-zone fraction, negative-surviving fraction ($\mathrm{IG}<{-}\delta$), and soft-clipping activation rate ($|\mathrm{IG}|>\eta$).}
\label{fig:ig_dist}
\end{figure}

Figure~\ref{fig:ig_dist} shows the histogram of raw (pre-stabilization) IG values across all search steps in a training batch at steps~50, 100, and~200 on Qwen2.5-3B-Base.

\par
Three trends confirm that each stabilization mechanism targets a meaningful subset of the data. 
First, the dead zone (gray band) captures 36\% of search steps at step~50 and still 28\% at step~200, predominantly from model-proficient questions whose near-zero IG does not reflect retrieval quality (Cases~1 and~3 in \S\ref{sec:deadzone_cases}).
Second, values exceeding the soft-clipping threshold $\eta{=}3$ grow from 15\% to 20\% of steps as the policy learns to issue more effective queries, consistent with the rising $\mathrm{IG}^{(0)}$ curve in Figure~\ref{fig:ig_by_step}; since negative values rarely exceed $-3$ nats, the clipping acts almost exclusively on the positive tail.
Third, the distribution shifts rightward over training (mean $0.27 \to 0.58$ nats), while the negative tail ($\mathrm{IG} < -\delta$) remains roughly constant at ${\sim}27$\%, confirming that asymmetric scaling is necessary to prevent these persistent negative signals from inducing search avoidance (Appendix~\ref{app:lambda_searchfreq}).

\section{Cross-Protocol Comparison with IGPO}
\label{app:igpo_comparison}

As noted in Appendix~\ref{sec:related_work}, IGPO and IG-Search differ in retrieval environment (Google Search API vs.\ a fixed Wikipedia corpus with E5-base-v2), rollout budget ($G{=}16$, $T_{\max}{=}10$ vs.\ $G{=}5$, $T_{\max}{=}5$), and training data.
These differences preclude inclusion in our main results table.
However, since IGPO reports word-level F1, we can compute the same metric for IG-Search to enable a comparison at the 3B scale.

Table~\ref{tab:igpo_compare} reports IG-Search-Instruct evaluated under F1 alongside the IGPO (w/ F1+IG) numbers copied from their paper (Qwen2.5-3B-Instruct).
IG-Search achieves an average F1 of 0.518 versus IGPO's 0.489, a 2.9-point improvement obtained despite a weaker retriever and a smaller rollout budget.

\begin{table}[h]
\centering
\caption{Cross-protocol comparison with IGPO on Qwen2.5-3B-Instruct (F1). IGPO numbers are copied from their paper. IG-Search numbers are computed under the same F1 metric.}
\label{tab:igpo_compare}
\small
\setlength{\tabcolsep}{3.5pt}
\begin{tabular}{l ccc cccc c}
\toprule
\multirow{2}{*}{\textbf{Method}} & \multicolumn{3}{c}{\textbf{Single-Hop QA}} & \multicolumn{4}{c}{\textbf{Multi-Hop QA}} & \multirow{2}{*}{\textbf{Avg.}} \\
\cmidrule(lr){2-4}\cmidrule(lr){5-8}
~ & NQ & TriviaQA & PopQA & HotpotQA & 2Wiki & Musique & Bamboogle & ~ \\
\midrule
IGPO (w/ F1+IG) & 0.419 & 0.692 & 0.490 & 0.478 & \textbf{0.514} & 0.248 & \textbf{0.584} & 0.489 \\
\rowcolor{cyan!10}
IG-Search-Instruct & \textbf{0.542} & \textbf{0.692} & \textbf{0.501} & \textbf{0.551} & 0.507 & \textbf{0.276} & 0.554 & \textbf{0.518} \\
\bottomrule
\end{tabular}
\end{table}

\section{Example of IG Computation on a Multi-Hop Question}
\label{app:ig_example}
\par
As described in \S\ref{sec:ig}, the step-level Information Gain is defined as the difference in the model's confidence in generating the gold answer when conditioned on the real retrieved context relative to a counterfactual baseline of random documents. 
To concretely illustrate this process, we provide a detailed illustration of a two-hop reasoning trajectory in Figure~\ref{fig:ig_example}. 

\par
This example demonstrates how the counterfactual contexts are constructed at each step while preserving the history of previous steps. 
Furthermore, it illustrates how the resulting IG values naturally capture the marginal informational contribution of successive retrieval calls: the first search introduces the novel gold answer and receives a high reward, whereas the second search provides only additional evidence, resulting in a positive but much smaller marginal gain.

\begin{figure}[htbp]
\centering
\small
\setlength{\tabcolsep}{3pt}
\begin{tabular}{p{0.96\textwidth}}
\toprule
\textbf{Question:} Who is the father of the father of modern experimental psychology? \\
\textbf{Gold Answer:} \textcolor{goldanswer}{\textbf{Maximilian Wundt}} \\
\midrule
\textbf{Step 1} \\[3pt]
\texttt{<think>} I need to find who is considered the father of modern experimental psychology. \texttt{</think>} \\
\texttt{<search>} father of modern experimental psychology \texttt{</search>} \\
\texttt{<documents>} Wilhelm Wundt is widely regarded as the father of experimental psychology. He was born on 16 August 1832 to parents \textcolor{goldanswer}{\textbf{Maximilian Wundt}}\ldots \texttt{</documents>} \\
\texttt{<refine>} Wilhelm Wundt is the father of experimental psychology. \texttt{</refine>} \\[4pt]
\rowcolor{cyan!10}
\textit{\textbf{IG computation}:} $\mathcal{C}_1^{\text{real}}$ contains the actual documents about Wilhelm Wundt. Three counterfactual contexts $\mathcal{C}_1^{\text{rand},j}$ replace these with random (documents, refinement) pairs from \textcolor{red}{other questions in the batch}. \\[3pt]
\rowcolor{cyan!10}
$\log \pi_{\theta}(\text{Maximilian Wundt} \mid \mathcal{C}_1^{\text{real}}) = -2.07$, \quad $\frac{1}{N}\sum_j \log \pi_{\theta}(\text{Maximilian Wundt} \mid \mathcal{C}_1^{\text{rand},j}) = -3.59$ \\[2pt]
\rowcolor{cyan!10}
$\Rightarrow$ \textcolor{igpositive}{\textbf{IG$_1$ = 1.52}}. \\
\rowcolor{cyan!10}
The actual documents mention \textcolor{goldanswer}{Maximilian Wundt} by name, substantially improving confidence over \textcolor{red}{random documents from unrelated questions}. \\
\midrule
\textbf{Step 2} (conditioned on Step 1 with its original documents) \\[3pt]
\texttt{<think>} Now I need to find the father of Wilhelm Wundt. \texttt{</think>} \\
\texttt{<search>} Wilhelm Wundt father \texttt{</search>} \\
\texttt{<documents>} Wundt was born in Neckarau, Baden, the fourth child to parents \textcolor{goldanswer}{\textbf{Maximilian Wundt}}, a Lutheran minister\ldots \texttt{</documents>} \\
\texttt{<refine>} Wilhelm Wundt's father was \textcolor{goldanswer}{\textbf{Maximilian Wundt}}. \texttt{</refine>} \\[4pt]
\rowcolor{cyan!10}
\textit{\textbf{IG computation}:} $\mathcal{C}_2^{\text{real}}$ includes Step 1 (with its original documents) followed by Step 2's actual documents. The counterfactual contexts $\mathcal{C}_2^{\text{rand},j}$ keep Step 1 intact but replace only Step 2's (documents, refinement) pair with \textcolor{red}{random ones from other questions}. \\[3pt]
\rowcolor{cyan!10}
$\log \pi_{\theta}(\text{Maximilian Wundt} \mid \mathcal{C}_2^{\text{real}}) = -0.41$, \quad $\frac{1}{N}\sum_j \log \pi_{\theta}(\text{Maximilian Wundt} \mid \mathcal{C}_2^{\text{rand},j}) = -1.15$ \\[2pt]
\rowcolor{cyan!10}
$\Rightarrow$ \textcolor{igpositive}{\textbf{IG$_2$ = 0.74}}. \\
\rowcolor{cyan!10}
Step 2 provides useful confirmation, but its marginal contribution is smaller because the \textcolor{goldanswer}{gold answer} already appeared in Step 1's documents. \\
\bottomrule
\end{tabular}
\captionof{figure}{Step-level IG computation on a two-hop question. Step~1 yields high IG because its documents contain the gold answer. $\mathcal{C}_2^{\text{rand}}$ keeps Step~1 completely and only replaces Step~2's (documents, refinement), so the counterfactual already carries Step~1's evidence; Step~2's IG therefore measures the \emph{marginal} value of the second retrieval.}
\label{fig:ig_example}
\end{figure}



\end{document}